\documentclass{article}

\usepackage[english]{babel}

\usepackage[letterpaper,top=2cm,bottom=2cm,left=3cm,right=3cm,marginparwidth=1.75cm]{geometry}

\usepackage{amsmath,amsfonts}
\usepackage{graphicx}
\usepackage{algorithmic}
\usepackage{array}
\usepackage[colorlinks=true, allcolors=blue]{hyperref}
\usepackage{cite}
\usepackage{authblk}

\usepackage{textcomp}
\usepackage{stfloats}
\usepackage{url}
\usepackage{verbatim}
\usepackage{caption}   
\usepackage{bm} 
\usepackage{xcolor}
\usepackage{balance}

\title{Haptic Perception via the Dynamics of Flexible Body Inspired by an Ostrich's Neck}
\author{Kazashi Nakano*,Katsuma Inoue,Yasuo Kuniyoshi,Kohei Nakajima}

\date{}

\begin{document}
\maketitle

\section*{Keywords}
{Biologically Inspired Robotics, Morphological Computation, Soft Robotics, Physical Reservoir Computing, Embodied Intelligence}

\begin{abstract}
In biological systems, both skin sensitivity and body flexibility play crucial roles in haptic perception.
Fully soft robots often suffer from structural fragility and delayed sensory processing, limiting their practical functionality.
The musculoskeletal system combines the adaptability of soft materials with the durability of rigid-body robots.
It also leverages morphological computation, where the morphological structures contribute to information processing, for dynamic and adaptive behaviors. 
This study focuses on the pecking behaviors of birds, which enables precise haptic perception through the musculoskeletal system of their flexible neck.
Physical reservoir computing is applied to flexible structures inspired by an ostrich neck to analyze the relationship between haptic perception and physical characteristics.
Experiments with both a physical robot and simulations reveal that, with appropriate viscoelasticity, the flexible structure can discriminate object softness and retain that information through behavior. 
Drawing on these findings and anatomical insights from the ostrich neck, a haptic perception system is proposed that exhibits both separability and behavioral memory in flexible structures, enabling rapid learning and real-time inference.
The results demonstrate that through the dynamics of flexible structures, diverse functions can emerge beyond their original design as manipulators.
\end{abstract}

\section{Introduction}
\label{sec: Introduction}

Haptic perception is the only sense among the five that arises from dynamic interactions with the environment\cite{gibson1966senses}.
The flexibility of the skin facilitates safe interactions and enables the encoding of information related to object properties, such as friction\cite{johansson2009coding}.
Although many studies have focused on flexible skin, researchers in ecological psychology have focused on the flexible body as the medium of touch\cite{turvey2014medium}.
For example, when a stick held in our hand is swung, its length can be perceived even if the stick is not visible.
In this case, information about the environment is obtained from the relationship between the active forces imposed by the muscles, the resistance provided by the object, and the resulting motions\cite{amazeen1996weight}.
By reflecting this environmental information throughout a flexible body, the precision and redundancy of perception are improved\cite{turvey2014medium,Mangalam2020}.
Haptic perception from mechanoreceptors in the skin is called cutaneous sensing, while that from mechanoreceptors in muscles, tendons, and joints is called kinesthetic sensing\cite{lederman2009haptic, Proske2012}.
\par Inspired by the haptic perception mechanisms of biological systems, soft robots with both cutaneous sensation and kinesthetic sensation have been developed\cite{Dahiya2010,Zhuofan2023}.
The deformation of flexible skin during contact converts external stimuli into large-amplitude derivative quantities, enabling the detection of fine details such as surface texture\cite{viry2014flexible,Cutkosky1993,howe1993tactile}.
The propagation of external information through a soft body has been demonstrated to improve the accuracy and redundancy of perception in soft robots\cite{thuruthel2019soft}.
However, soft robots with flexibility throughout their bodies face challenges in body design and sensory information processing.
One major issue in body design is the significant temporal variation caused by material degradation or malfunction, especially in robots made of soft materials such as silicone or polyurethane\cite{terryn2021review}.
Furthermore, the motion of soft robots relies primarily on their own elasticity, its ability to perform dynamic tasks is significantly limited by the slow rate of deformation\cite{pal2021exploiting}.
Sensory information processing also presents challenges.
When soft robots interact with the environment, both the robot and the environment typically exhibit dynamics, as in invasive surgery and quality assurance in the food industry\cite{Sornkarn2016}.
In this case, conventional force sensors suppress their own dynamics and assume a constant indentation depth in the environment, while soft robots exhibit nonlinear self-deformation that is difficult to model\cite{Konstantinova2014}.
Therefore, learning-based approaches such as Long Short-Term Memory (LSTM)\cite{Truby2020}, Convolutional Neural Network (CNN)\cite{wang2020real}, and autoencoders\cite{Soter2018} are promising solutions for learning complex self-deformations\cite{Benjamin2020,Zhang2023}.
However, the time delay in inference caused by nonlinear models remains a problem, and improving the time delay in self-deformation calculations requires a heavy computational load\cite{kim2021review}.
\par  The musculoskeletal system found in many organisms can achieve a balance between the continuous deformability inherent in soft robots and the robustness, high output and high precision characteristics of rigid robots\cite{Niikura2022}.
In the musculoskeletal system, constraints due to morphological structures such as joint structure and muscle-tendon arrangements can facilitate some intelligent behaviors.
For example, the viscoelasticity and anisotropic compliance inherent in the musculoskeletal system were utilized to perform complex and dynamic tasks, such as identifying objects and opening doors, using simple control inputs\cite{Hosoda2012}.
In this case, some of the calculations necessary for control are outsourced to the physical body itself\cite{pfeifer2014cognition}.
This framework is referred to as morphological computation\cite{pfeifer2006body}, and it has been shown that utilizing it enables intelligent behaviors such as gait, with fewer computational resources\cite{Fuchslin2013,mcgeer1990passive}.
We consider the use of the musculoskeletal system and its dynamics to be one of the most promising approaches to take.
Therefore, we focus on the pecking behavior of birds, which achieves precise tactile sensing using the flexible musculoskeletal system of the neck\cite{ziolkowski2022tactile}.
As shown in \textbf{Figure \ref{fig: Physical_reservoir}(a)}, we dissected an ostrich's neck and identified the muscle-tendon arrangement for its typical pecking behavior, and developed a flexible robot capable of performing pecking behaviors similar to those of an ostrich in our previous researches\cite{Mochiyama2022,nakano2023robostrich}.
(Supporting information video 1 shows the behavior of RobOstrich in detail.)
Furthermore, we confirmed that the robot exhibited a variety of dynamics (i.e., time series of joint angles) in response to different external forces, due to its passivity and multi-degree-of-freedom characteristics\cite{nakano2023kinesthetic}.
However, the behaviors were highly sensitive to physical characteristics.
In morphological computation, it has been reported that input and body parameters have a significant impact on the resulting behaviors and tactile sensation\cite{terajima2022,Sornkarn2016,Herzig2018}.
A comprehensive understanding of the relationship between input parameters, physical characteristics, and haptic perception is required.
\par  Therefore, the purpose of this study was to construct a haptic perception system that utilizes the dynamics of the avian musculoskeletal system and examine the effect of input parameters and body characteristics on haptic perception.
The focus is primarily on the body's viscoelasticity as a physical characteristic because, as pointed out in ecological psychology, connective tissues (muscles, tendons, and ligaments) contribute to haptic information processing as mechanical networks, much like neural networks\cite{turvey2014medium}.
Within this mechanical network, variations in elasticity affect the speed and location of force propagation within the body, whereas variations in viscosity alter the characteristics of the body's response to external forces.
The contributions of this paper relate to haptic perception through the dynamics of a biologically inspired musculoskeletal structure, specifically including the following.

\begin{itemize}
\item Experiments using both a physical robot and simulations demonstrate that the dynamics of a flexible body can amplify subtle differences in external forces---termed \textit{separability}---and preserve them as behavioral patterns---termed \textit{haptic memory}.
\item Detailed time-series analysis revealed the conditions of input and physical parameters that enhance separability and haptic memory.
 \item We evaluated the effect of heterogeneity in body viscoelasticity, resulting from morphological structures such as ligament and muscle arrangements, on both separability and haptic memory.
\item The system leverages physical reservoir computing for seamless data acquisition and learning, completing the training in approximately 25 seconds. It further enables real-time inference of the softness of collided objects.
\item We demonstrate that this haptic memory allows for state-dependent outputs, like a sequential circuit, and can reduce the computational load in sensor data processing.
\end{itemize}

These findings indicate that leveraging the inherent dynamics of a flexible body can lead to emergent functionalities beyond the original design intentions as a manipulator, highlighting their potential in biologically inspired soft robotic systems.

\section{Results}

\subsection{Physical Reservoir Computing}
\label{sec: Physical Reservoir Computing}
Morphological computation views the complex dynamics of flexible structures not only as a control problem but also as a computational resource\cite{pfeifer2006body}.
This computational capability can be formulated as the ability to emulate the mapping from input to output when a static and linear readout is added to a morphology such as a mass-spring model\cite{hauser2011towards}.
This method is commonly referred to as physical reservoir computing\cite{nakajima2020physical}.
In this approach, physical systems, such as the body of a soft robot, are regarded a recurrent neural network with fixed weights.
By training the readout layer, it is possible to emulate desired functions such as sensing and control\cite{Zhao2013,Wang2023}.
In physical reservoir computing, two inherent capabilities of physical systems are leveraged: the ability to non-linearly map low-dimensional inputs to a high-dimensional space, similar to kernels in machine learning, and the ability to temporarily retain the influence of recent input sequences\cite{Maass2002}.
Previous studies have thoroughly analyzed the positional distribution of computational capabilities that originate from the non-linearity and memory inherent in the body dynamics of soft robots\cite{nakajima2013soft}.
In the context of haptic perception in a flexible body, methods have also been developed to quantify and analyze the spatio-temporal propagation of information related to contact in a flexible body\cite{nakajima2015measuring}.
These studies show that the framework of physical reservoir computing and its analysis is useful to quantify the contribution of body parameters, morphology, and behavior to perception\cite{Luís2019}.

\subsection{Task Definition}

\subsubsection{Insights from Biology}
\begin{figure}[h]
\centering
  \includegraphics[width=0.9\linewidth]{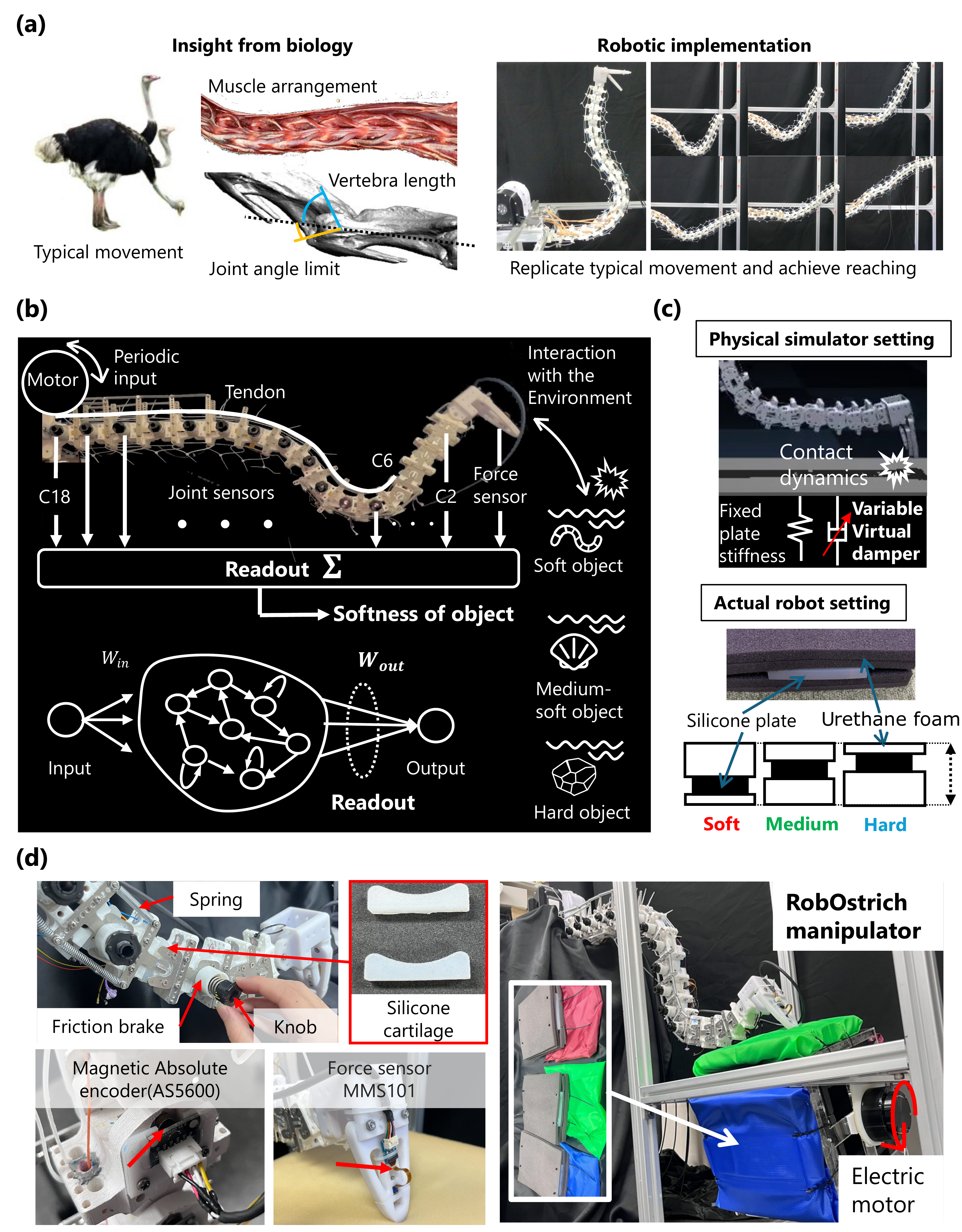}
  \caption{
    Overviews haptic perception via the dynamics of bio-inspired flexible manipulator:
    (a) RobOstrich: A tendon-driven flexible manipulator based on the anatomy of an ostrich's neck
    (b) haptic perception through physical reservoir computing utilizing the dynamics of RobOstrich
    (c) dynamic properties of the external environment in physical simulations and actual robot experiments
    (d) experimental setup in actual robot
    }

  \label{fig: Physical_reservoir}
\end{figure}

To investigate the relationship between physical characteristics and haptic perception, we focused on the pecking behavior of birds.
Birds have flexible necks with around 20 cervical vertebrae and repeatedly peck at the ground\cite{Marek2021}.
Some birds can detect prey hidden in the ground without relying on their eyesight, sensing the vibrations and the difference in pressure when they peck at the ground\cite{cunningham2007new,Cunningham2010Remote}.
Despite the fact that the beak with the mechanoreceptors is at the tip of a self-deforming flexible neck, some kinds of birds achieve an accurate perception that can distinguish between food and stones\cite{cunningham2007new}.
The pecking movement is characterized by high-speed movement, with a shorter time scale of the electromyography peak compared to other behaviors\cite{van2001control}.
Furthermore, certain species of birds close their eyes during the pecking movement\cite{Zweers1994}.
This suggests that the motor commands of cephalized control in the hierarchical structure of the sensory-motor system\cite{Ijspeert2023} are feedforward.
Based on these findings, we defined a haptic perception task in which the reaction during a collision was obtained and the softness of the collided object was classified using physical reservoir computing.
\textbf{Figure \ref{fig: Physical_reservoir}(b)} shows the concept of this task.
In this task, the RobOstrich manipulator that we previously developed repeatedly collides with the target object\cite{nakano2023robostrich}.
RobOstrich is a tendon-driven multi-degree-of-freedom passive joint structure with 17 degrees of freedom in the sagittal plane.
Here, the dynamics of RobOstrich were considered as a physical reservoir, and time series data of joint angles (17 elements) and beak forces (2 elements: tangential and vertical) were obtained.
Softmax regression (detailed in Section \ref{sec: Softmax Classifiers}) was applied as a readout to the collected time series data (both joints and force) to classify the softness of the target object.
This setup is similar to that of the work\cite{horii2021physical}, but differs in that the body and environment periodically detach during pecking movement.
The input is the displacement for pulling the tendon by the motor located at the base.
A sinusoidal forced displacement is applied to the end point of the tendon as the input.
\begin{equation}
u(t) = AR \cos\left(\frac{2\pi}{T}t\right) + Dr
\end{equation}
Here, $u$ is the position of the endpoint [m], $A$ is the amplitude of the motor's rotational angle [deg] that winds the tendon, $R$ is the diameter [m] of the motor pulley, $T$ is the period [s], and $Dr$ is the drift [m].
The endpoint of the tendon is connected to the sixth cervical vertebra (C6) of the head, and the segment closer to the head is an elastic joint that is not connected to any input(Figure \ref{fig: Physical_reservoir}(b)).
This tendon mimics the tendon of the particularly well-developed muscle in the neck of the ostrich\cite{nakano2023robostrich}.
The transmission of force in the tendon-driven system is anisotropic; while it can transmit tension by pulling the flexible structure, rapidly reversing the tension allows the tendon to slack, enabling movement that follows the inherent dynamics of the flexible structure.
Constraints in each joints in RobOstrich allow for natural pecking movements even with simple control inputs.
The collision with the environment during the pecking motion itself also has dynamics.
In this task, the dynamics of the multi-degree-of-freedom flexible body are utilized to classify the dynamic properties (damping coefficient) inherent in the environment through the dynamics of this collision.
In the physical simulator, differences in learning performance due to variations in input and body parameters are comprehensively evaluated.
In actual robot experiments, the insights from the simulation are validated, and the potential applications of the proposed system are discussed.
The simulator experiment is a 9-class classification (upper figure of \textbf{Figure \ref{fig: Physical_reservoir}(c)}), while the actual robot experiment is a 3-class classification (lower figure of Figure \ref{fig: Physical_reservoir}(c)).
\textbf{Figure \ref{fig: Physical_reservoir}(d)} shows the method for changing the robot's physical characteristics, the sensors and experimental setup.
The details of the physical simulator and the actual robot experiments are provided in Section \ref{sec: Experimental Setup in Physical Simulator} and \ref{sec: Experimental Setup in Actual Robot}.

\subsubsection{Accuracy Curve}
\label{sec: Accuracy Curve}
\begin{figure}
\centering
  \includegraphics[width=0.8\linewidth]{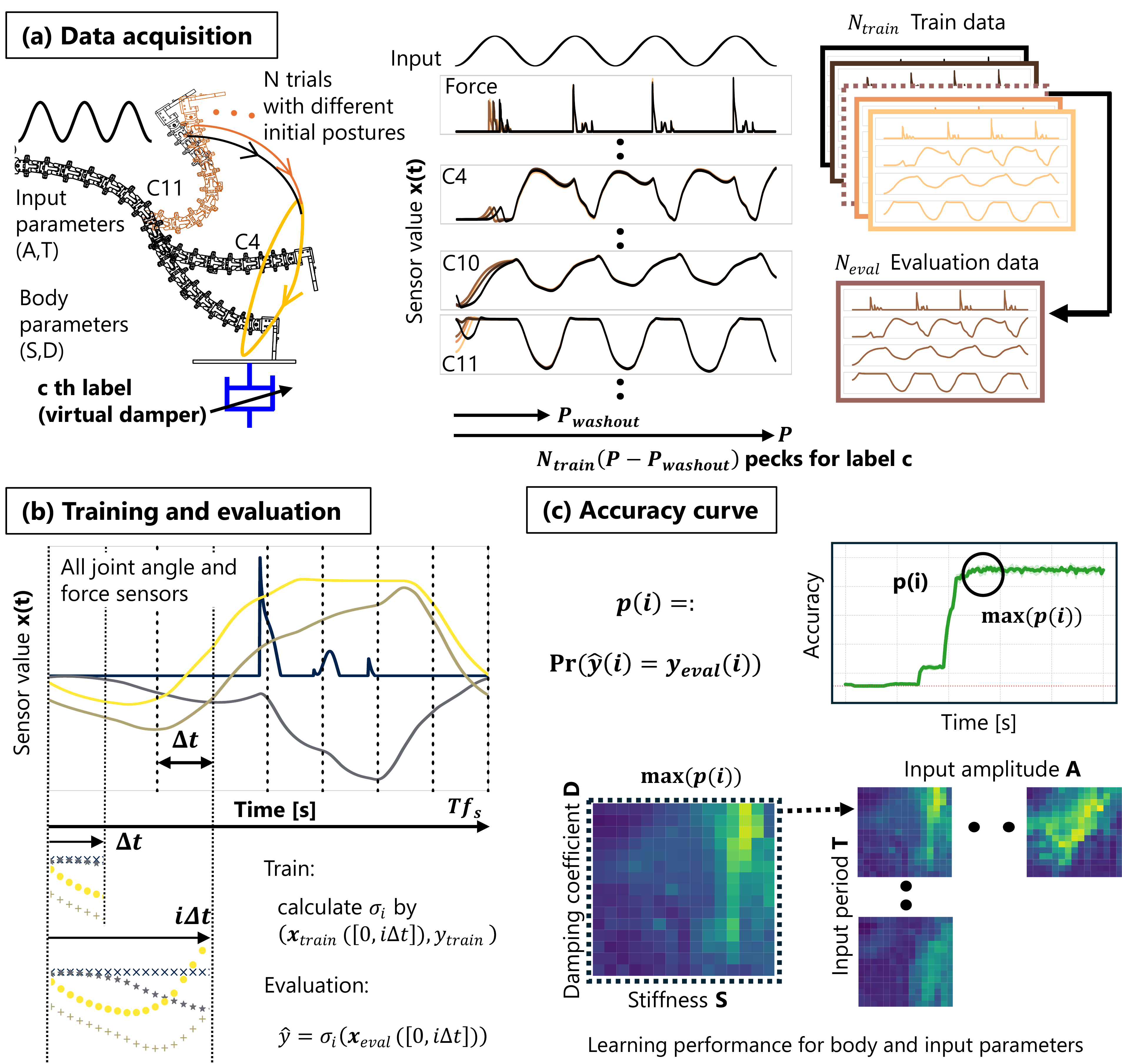}
  \caption{
   Definition of the accuracy curve and exploration of the distribution of learning performance based on the accuracy curve:
   (a) data acquisition method
   (b) time-multiplexing in training and evaluation
   (c) exploring learning performance with respect to input and physical parameters
    }

  \label{fig: get_data_and_learning}
\end{figure}

o evaluate the computational capacity of RobOstrich dynamics and to assess when this capability is observed, we defined the accuracy curve using time multiplexing \cite{Nakajima2013multiplex}.
First, as shown in \textbf{Figure \ref{fig: get_data_and_learning}(a)}, an input with amplitude $A$ and period $T$ was applied to a body with viscoelasticity $(S, D)$.
Then, sensor time series were acquired for $C$ types of plates with damping coefficients, from $P$ periods of pecking motions, starting from $N$ initial states of the neck ($C, N, P \in \mathbb{N}$).
The washout phase to reach the steady state of the reservoir was set to $P_{washout}$ periods, and the remaining $P-P_{washout}$ periods of the sensor time series data were used.
Among the time series with $N$ types of initial states, $N_{train}$ types were selected as training data and the $N_{eval}$ types were evaluation data.
Here, for each label, $N_{train}(P - P_{washout})$ periods of training data were obtained.
In the case of a pecking period $T$ and a sampling rate $f_s$, one cycle of time series data contains $Tf_s$ data points.
Each sensor data for one period is divided into intervals of $\Delta{t}$ ($\Delta{t} \in \mathbb{N}$), and the data from these time intervals were accumulated to apply softmax regression as shown in \textbf{Figure \ref{fig: get_data_and_learning}(b)}.
Focusing on the $i$-th interval, we can obtain the model $\sigma_i$ at that point in time using the accumulated training data $(\mathbf{x}_{train}([0, i\Delta t]),y_{train}(i))$ as follows:

\begin{equation}
\label{eq: from_zero_to_it}
\hat y(i) = \sigma_i(\mathbf{x}_{eval}([0, i\Delta t]))
\end{equation}

By calculating the provability of instances where $\hat y(i) = y_{eval}(i)$ among the $P - P_{washout}$ trials out of the $N$ experiments, we obtain the time variation of the accuracy $p(i)$.
Thus, $p(i)$ can be formulated as follows:
\begin{equation}
p(i) = \frac{1}{N_{eval}(P - P_{\text{washout}})} \sum_{n=1}^{N_{eval}} \sum_{p=1}^{(P - P_{\text{washout}})}\delta(\hat{y}(i)_{n,p}, y_{\text{eval}}(i)_{n,p})
\end{equation}

Here, $\delta(\hat{y}(j), y_{\text{eval}}(j))$ is the Kronecker delta function, which returns 1 if the estimate $\hat{y}(j)$ matches the evaluation value $y_{\text{eval}}(j)$ and 0 otherwise.
As shown in \textbf{Figure \ref{fig: get_data_and_learning}(c)}, we refer to the plot of $p(i)$ against $i$ as the accuracy curve.
In this $p(i)$, the mean and standard deviation are calculated based on the combinations of $N_{eval}$ evaluation data selected from the different initial states $N$.
In other words, we evaluate the generalization performance of the classification task for the object's damping coefficient with respect to the initial state.
Furthermore, by changing the viscoelasticity of the body in 225 patterns and plotting the maximum values of the accuracy curve represented by Equation (\ref{eq: max_pi}) as a heat map, the distribution of the computational capabilities of the body with respect to its viscoelasticity is determined.

\begin{equation}
\label{eq: max_pi}
Accuracy = \max_{0 \leq i \leq L} p(i), \quad \text{where } L = \frac{Tf_s}{\Delta t}
\end{equation}

The range for plotting viscosity and elasticity to plot performance is based on prior research on the viscoelasticity of human knees\cite{Kooij2022}, with stiffness set between 0.1 and 17.5 Nm/rad and the damping coefficient set between 0.02 and 1.5 Nms/rad.
Previous research investigating the viscoelastic properties of human knee joints during walking has indicated that in the swing phase, the stiffness is approximately 0.0-3.5 Nm/rad and the damping coefficient is 0.02-0.14 Nms/rad. In the stance phase, the stiffness ranges from 15 to 20 Nm/rad and the damping coefficient ranges from 1.0 to 2.0 Nms/rad\cite{Kooij2022}.
Furthermore, by determining this distribution for each input parameter, the relationship between learning performance and input and body parameters is acquired.

\subsection{Behaviors that Enable Haptic Perception}
\label{subsec: Relationship between haptic perception and behaviors}

\begin{figure}
  \centering
  \includegraphics[width=0.93\linewidth]{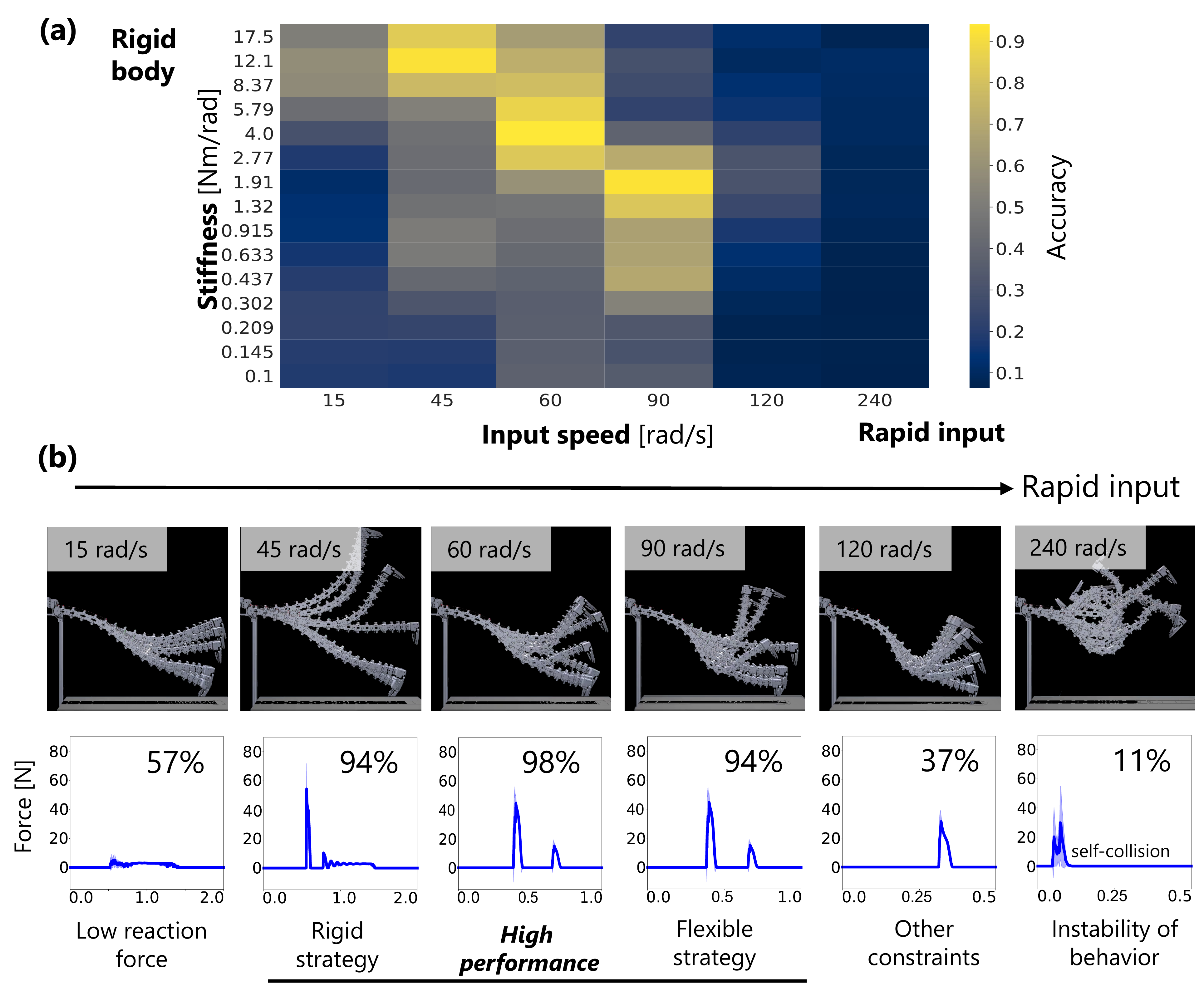}
  \caption{Distribution of learning performance with respect to input and body parameters: 
  (a) relationship between input and body parameters under high performance conditions
  (b) behaviors and reaction force under high performance conditions
  }
  \label{fig: heat maps_and_behaviors}
\end{figure}

\textbf{Figure \ref{fig: heat maps_and_behaviors}(a)} shows a heat map that illustrates the relationship between body stiffness $S$ and input speed $A/T$ at the maximum value of learning performance.
\textbf{Figure \ref{fig: heat maps_and_behaviors}(b)} shows the behaviors and the time series of the reaction force at the maximum learning performance for each input speed.
The supporting information provides the learning performance for all patterns (12 patterns for input parameters and 225 patterns for body parameters), along with the corresponding behavior and time series data.
Figure \ref{fig: heat maps_and_behaviors} (a) shows that learning performance is low for inputs that are either too rapid (240 rad/s) or too slow (15 rad/s).
As shown in Figure \ref{fig: heat maps_and_behaviors} (b), at 240 rad/s, the input exceeds the viscoelastic capacity of the body, disrupting stable pecking.
Additionally, at 15 rad/s, the generated reaction force is insufficient to obtain meaningful information about the pecked object.
These observations indicate that physical conditions, such as the inability to maintain stable pecking, and informational conditions, such as insufficient reaction force to extract information from the environment, both play crucial roles in enabling learning.
In these examples, the conditions that reduce learning performance are clear.
Notably, even under seemingly favorable conditions such as 120 rad/s—with stable pecking and stronger reaction force than at 15 rad/s—the learning performance remains lower, suggesting more subtle underlying physical and informational conditions.
Even in cases where the possibility of learning is not clear from the behavior, physical and informational conditions are likely to exist.
Focusing on conditions with high learning performance (45-90 rad/s), as the input speed increases, the value of body stiffness that corresponds to the maximum value of the heat map decreases.
In other words, there are strategies that involve driving a flexible body at high speed or a rigid body at low speed.
The diversity of these strategies is thought to contribute to adaptability, allowing the behavior to be modulated according to different environmental conditions or task demands\cite{warren2006dynamics}.
The behavior under each condition is shown in the supporting information video 2.
\par Overall, we can conclude that 1)a flexible body driven by a tendon can exhibit various behaviors depending on the input and body parameters, but there are physical and informational conditions to achieve haptic perception; and 2) there are multiple solutions that maximize learning performance in haptic perception, and the input and body parameters in these solutions are interdependent.
These conditions are analyzed in detail in Sections \ref{subsec: echo state property} and \ref{subsec: Separability}.
In Section \ref{subsec: haptic memory}, we discuss the memory capability that becomes available when a flexible body is used at high speed.
Given that the pecking period of the ostrich is approximately 0.5 to 1.0 seconds, the following discussion focuses on the conditions of the rapid input parameters ($A$ = 90 deg and $T$ = 1.0 s).

\subsection{Echo State Property}

\label{subsec: echo state property}
This section clarifies the conditions of learning performance from the perspective of physical reservoir computing.
To enable physical reservoir computing, the physical reservoir should satisfy the echo state property (ESP)\cite{jaeger2001echo}.
The ESP is an asymptotic stability property with respect to different initial values of the reservoir state vector.
In this system, if the ESP does not hold, learning performance is expected to deteriorate significantly.
The time evolution of the physical reservoir state $\mathbf{x}$ can be written as follows, where $\mathbf{u}$ is the input and $\mathbf{F}$ is the state transition function:

\begin{equation}
\label{eq: state_dev}
\mathbf{x}(n+1) = F(\mathbf{u}(n),\mathbf{x}(n))
\end{equation}

Additionally, the resulting reservoir state vector from the initial state $x_0$ and the input time series $\mathbf{U}(n)=[\mathbf{u}(0),\mathbf{u}(1),,,\mathbf{u}(n)]$ can be expressed as:
\begin{equation}
\mathbf{x}(n+1) = \hat{F}(\mathbf{U}(n),\mathbf{x(0)})
\end{equation}
Here, for the reservoir to satisfy the ESP, the following condition must be satisfied for any initial states $\mathbf{x_0}$,$\mathbf{z_0}$, and any input time series $\mathbf{U}(n)$\cite{gallicchio2019chasing}:

\begin{equation}
\label{eq: esp}
\lim_{n \to \infty} \| \hat{F}(\mathbf{U}(n),\mathbf{x_0})- \hat{F}(\mathbf{U}(n),\mathbf{z_0})\|_2 =0
\end{equation}

This means that the distance between the reservoir state vectors, starting from different initial states, converges to 0.
Based on this, in this paper, we define the ESP index $I$ as follows:

\begin{equation}
\label{eq: esp_index}
\begin{aligned}
I &= \langle \overline{\bm{\sigma}(t)} \rangle \\
\bm{\sigma}(t) &= \frac{1}{N} \sum_{n=1}^{N} \left\|\mathbf{x}^n(t) - \frac{1}{N} \sum_{n=1}^{N} \mathbf{x}^n(t)\right\|
\end{aligned}
\end{equation}

\begin{figure}
\centering
  \includegraphics[width=\linewidth]{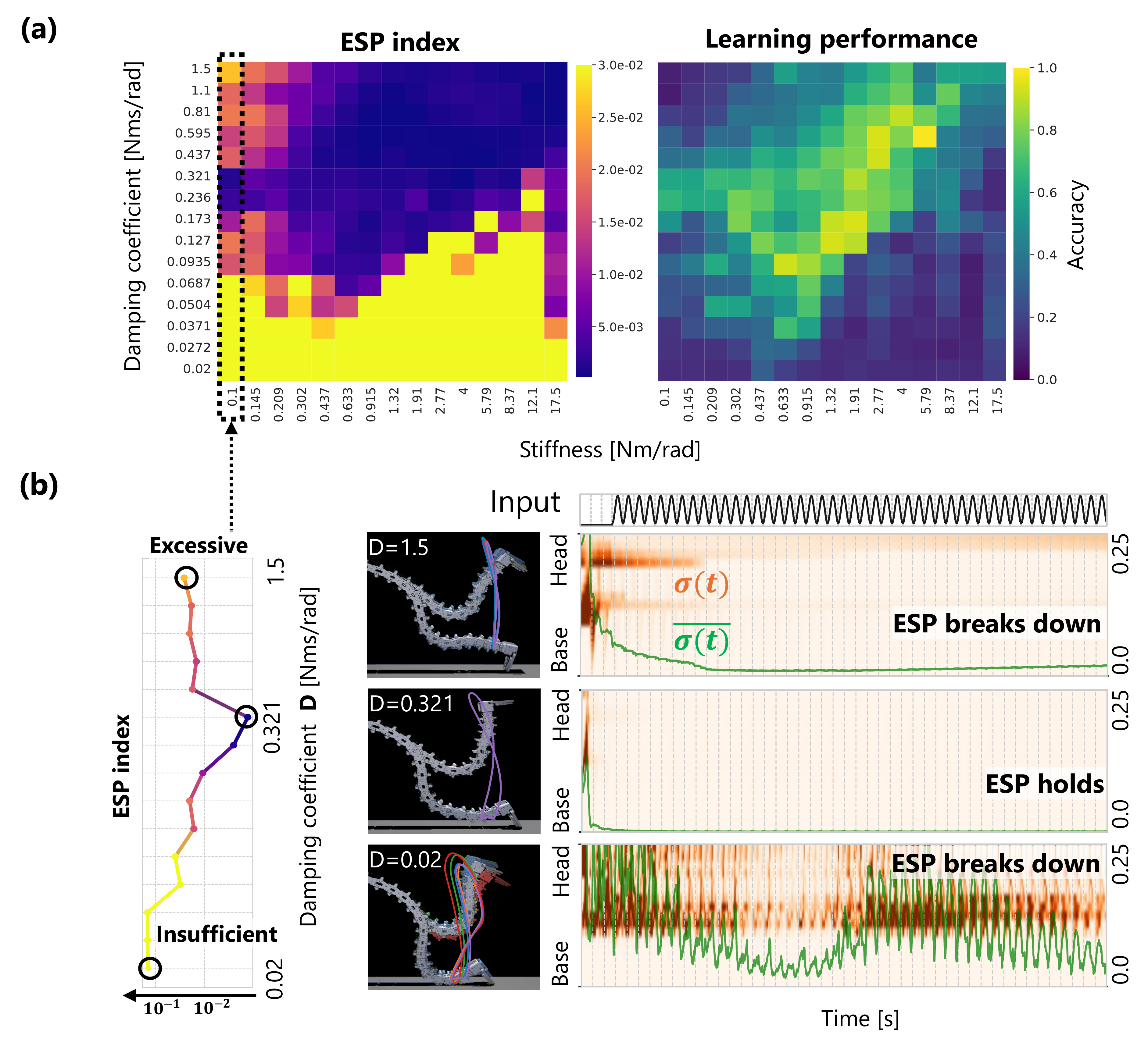}
  \caption{
  Behavior that enables learning and the corresponding viscoelasticity of the body:
  (a) comparison of heat maps for ESP index and learning performance
  (b) differences in behavior due to variations in initial conditions
  }
  \label{fig: ESP}
\end{figure}

Here, $\mathbf{X}^n \ (n=1,2,\dots,N)$ represents a time series of joint angles with $N$ different initial values.
In addition, $\langle \bullet \rangle$ represents the time average and $\overline{\bullet} $ represents the component average.
If these $I$ and $\bm{\sigma}(t)$ are close to 0, the ESP is satisfied.
The left heat map in \textbf{Figure \ref{fig: ESP}(a)} presents the ESP index $I$ with respect to the viscoelasticity of the body.
Compared to the learning performance shown on the right in Figure \ref{fig: ESP}(a), we can see that the region with high learning performance is generally within the area where ESP holds.
Therefore, the actual factors that decrease learning performance can be understood by examining the time series of joint angles for different initial states and the same input.
The left figure in \textbf{Figure \ref{fig: ESP}(b)} shows the postures and tip trajectories under conditions of excessive viscosity (where ESP breaks down), moderate viscosity (where ESP holds) and insufficient viscosity (where ESP breaks down), color-coded according to the differences in the initial states.
The right figure in Figure \ref{fig: ESP}(b) shows the time series of $\bm{\sigma}(t)$ (color map) and $\overline{\bm{\sigma}(t)}$ (green line) under each condition.
These figures demonstrate that, under excessive viscosity conditions, the $\sigma$ and $\overline{\sigma}$ values near the head do not reach zero due to the steady deviations in posture, even after repeated pecking.
This is likely due to the greater influence of non-linear internal forces, such as tendon friction, compared to the reaction forces.
Furthermore, in regions of insufficient viscosity, $\sigma$ does not reach zero at various positions of the body, resulting in a large deviation in the tip trajectory.
This indicates the instability of the behavior due to insufficient damping against the reaction force.
Even in situations where steady pecking seems to be achieved, the time series of joint angles can be unstable.
Therefore, we consider non-linear internal forces and the instability of behavior as physical conditions of learning performance.
For heat maps of the ESP with other input parameters, refer to the supporting information.

\subsection{Separability}
\label{subsec: Separability}
\begin{figure}
\centering
  \includegraphics[width=\linewidth]{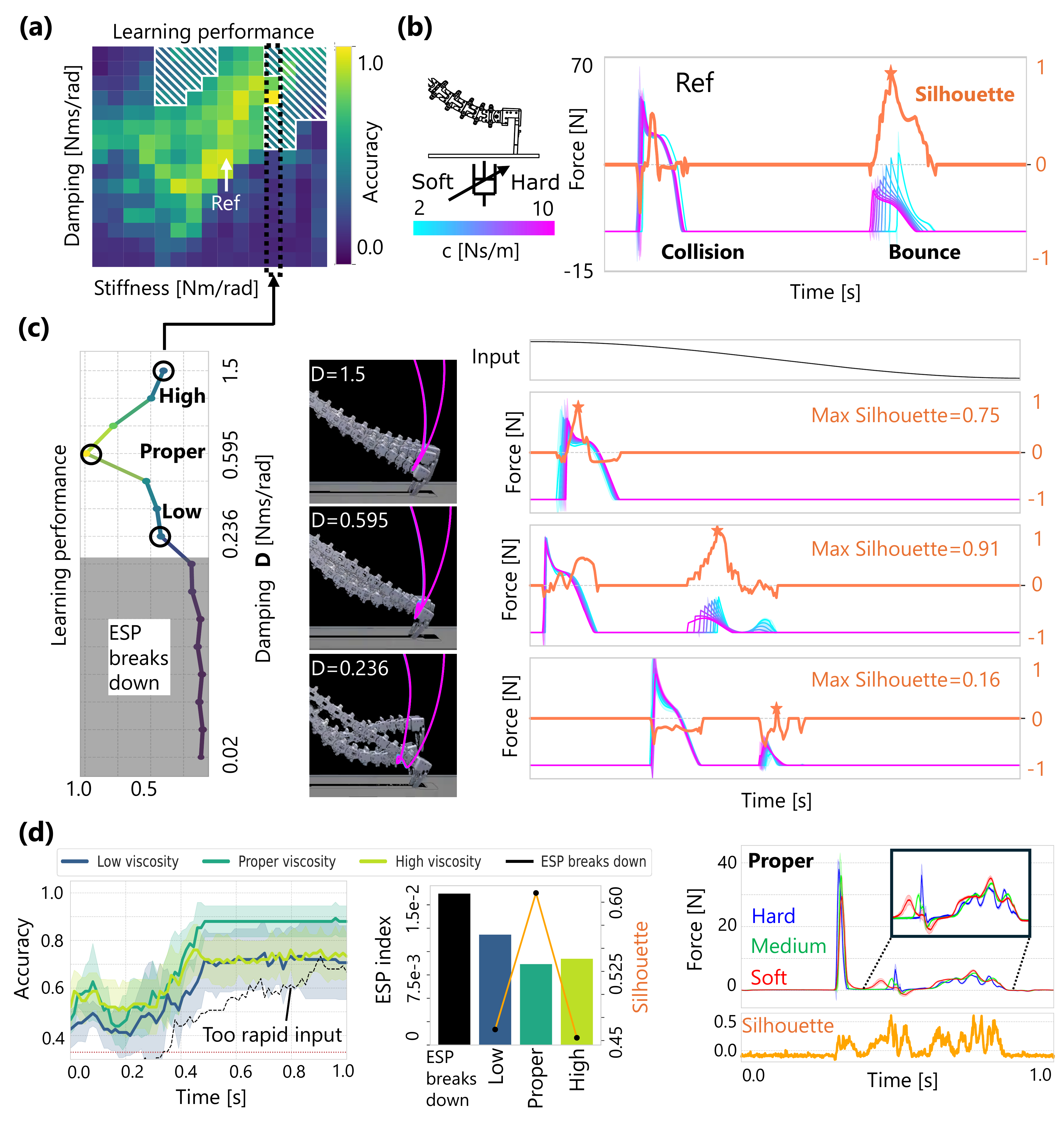}
  \caption{
  Adaptive behavior that amplifies the differences in reaction forces and the corresponding viscoelasticity of the body:
  (a) region of Low Learning Performance where ESP holds (White Frame)
  (b) separation of reaction force time series in response to differences in objects observed after the collision
  (c) relationship between physical characteristics and the time series of reaction force
  (d) changes in learning performance due to adjustment of body parameters in actual robot experiments
  }
  \label{fig: Separability}
\end{figure}

Even if ESP holds, it does not necessarily lead to high learning performance.
The regions enclosed by the white lines on the heat map of \textbf{Figure \ref{fig: Separability}(a)} represent areas where ESP holds, but learning performance remains low.
In this case, the ESP reflects the stability of RobOstrich behavior (joint angles).
Furthermore, analysis of the reaction force time series may provide insights into the conditions that enhance learning performance.
\textbf{Figure \ref{fig: Separability}(b)} shows the time series of the reaction force at the local maximum of the heat map.
The time series of reaction force are color-coded according to the dynamic properties of the object (damping coefficient), and when the colors are clearly separated, it indicates high learning performance.
To quantify this separation, we use the silhouette coefficient, which is described in detail in Section \ref{sec: Silhouette coefficient}.
The silhouette coefficient ranges from $[-1, 1]$. Values near 1 indicate well-separated clusters, 0 suggests overlapping clusters, and negative values imply misassignment.
The orange line in Figure \ref{fig: Separability}(b) shows the time series of the silhouette coefficient.
The star symbol represents the maximum value of the silhouette coefficient, indicating the moment when the separation is most distinct.
These plots in Figure \ref{fig: Separability}(b) indicate that, at the moment of collision, trajectory separation is unclear but becomes clear during the subsequent bounce.
This is likely because, at the moment of contact, the impulsive impact force dominates over the dynamic characteristics that should be classified.
\textbf{Figure \ref{fig: Separability}(c)} compares the reaction force time series with different body viscosities and their corresponding behaviors under the condition.
With higher viscosity, the amplitude of the rebound force decreases and eventually disappears, while with lower viscosity, the trajectory deviation increases, making the separation unclear.
Overall, a flexible body can exhibit behavior that steadily amplifies the difference in reaction forces during contact (separability), which is a condition to achieve high learning performance.
\par In the following, these findings are validated through actual robot experiments.
First, we compare the change in accuracy with respect to the input amplitude $A$.
A larger input amplitude corresponds to a rapid input speed.
\textbf{Figure \ref{fig: Separability}(d)} shows the accuracy curve.
The x-axis represents the time over one pecking cycle and the y-axis represents the accuracy.
The black line in the accuracy curve of Figure \ref{fig: Separability}(d) represents the accuracy under conditions with a relatively rapid input speed (e.g., $A/T$ = 80 deg/s), while the colored lines represent conditions with a slow input speed (e.g., $A/T$ = 20 deg/s).
The black line shows that, under the rapid input condition, the learning performance remains low, even when the object is periodically colliding with the system.
The center part of Figure \ref{fig: Separability}(d) shows the ESP index and the maximum values of the silhouette coefficient for each case.
When the input speed is high, the ESP index increases, indicating that the ESP does not hold.
In other words, by focusing on the reproducibility of the joint angle time series, we can determine the input parameters that enable learning.
Next, we examine accuracy changes by adjusting the viscosity with a friction brake, while keeping the input parameters fixed under conditions where the ESP holds.
As shown in the accuracy curve, learning performance is low when the body viscosity is either too high or too low.
Additionally, as shown in the center part of Figure \ref{fig: Separability}(d), the learning performance is highest under conditions where the silhouette coefficient are maximized.
This means that by focusing on the separation and deviation of rebound forces, we can estimate the learning performance.
The right part of Figure \ref{fig: Separability}(d) shows the time series of the reactive force and the silhouette coefficient for proper body viscoelasticity.
This figure illustrates that, in actual robot experiments, the separation of the trajectory occurs during the bounce that follows the impact force.
By focusing on the reproducibility of body dynamics (Echo State Property) and the separability and deviation of the reactive force time series, we can set appropriate input and body parameters before training and evaluation.
The data from these actual robot experiments are shown in the supporting information.

\subsection{Haptic Memory}
\label{subsec: haptic memory}

\begin{figure}
\centering
  \includegraphics[width=0.9\linewidth]{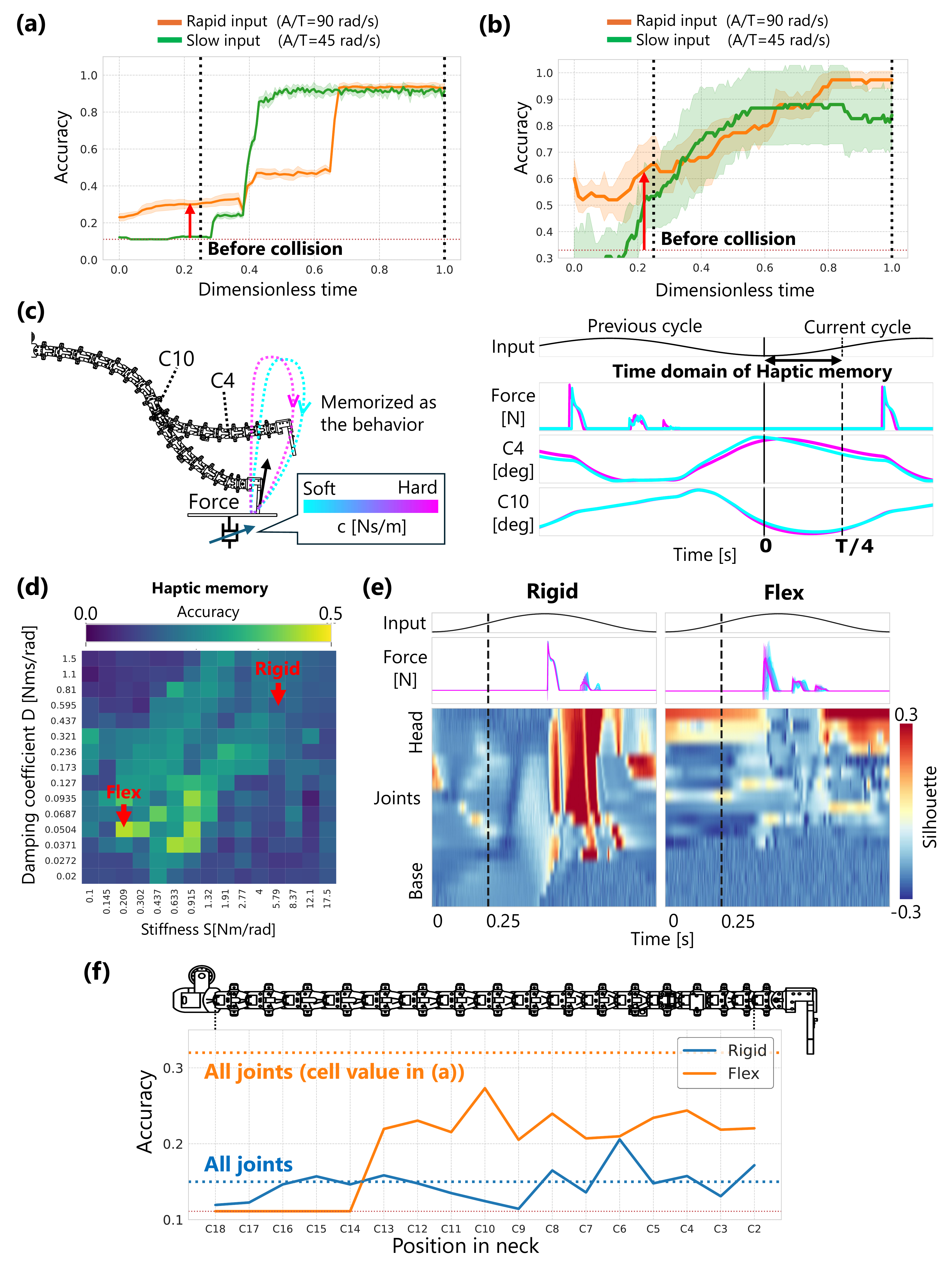}
  
  \caption{
    Introduction and analysis of haptic memory:
    (a) Learning performance existing in the dynamics before the collision
    (b) Verification of learning performance before the collision on an actual robot
    (c) Differences in behavior that are retained long term due to differences in objects
    (d) distribution of haptic memory with respect to body parameters
    (e) time series of reaction force and silhouette coefficient of joint angles
    (f) distribution of haptic memory in the body
   }
  \label{fig: haptic_memory}
  
\end{figure}

\textbf{Figure \ref{fig: haptic_memory}(a)} presents the accuracy curves under the conditions of $A/T=$ 90 rad/s (rapid input) and 45 rad/s (slow input) in the simulator.
The x-axis represents the dimensionless time over one pecking cycle and the y-axis represents the accuracy.
In each condition, an accuracy of about 90 percent is finally achieved with appropriate body viscoelasticity.
Note that under rapid input conditions, low body viscoelasticity (flexible strategy) is required, while under slow input conditions, high body viscoelasticity (rigid strategy) is needed.
In rapid input, the accuracy curve exceeds the chance rate (indicated by red dashed lines) for the classification even before collision (indicated by the red arrow).
As shown in \textbf{Figure \ref{fig: haptic_memory}(b)}, the existence of learning performance even before collision is confirmed in actual robot experiments.
The details of the experiment are shown in the supporting information video 3.
We consider that the existence of learning performance before collision is due to the information from the previous cycle being ``memorized as the behavior" during dynamic periodic movements, as shown in \textbf{Figure \ref{fig: haptic_memory}(c)}.
In biological systems, haptic information is stored hierarchically according to its forgetting time, and sensory information temporarily stored in the sensory nervous system is called haptic memory\cite{carlson2007physiology,shih2009evidence}.
The hierarchical structure of the memory mechanisms allows more attention to be focused on more critical information, thereby improving adaptability\cite{Changjin2020}.
Therefore, we focus on this ``memory as behavior" and define haptic memory as learning performance using only the data before the collision ($0 \leq t \leq T/4 [s]$):

\begin{equation}
\label{eq: memory}
Accuracy = \frac{1}{L} \sum_{i=0}^{L} p(i), \quad \text{where} \quad L = \frac{T f_s}{4 \Delta t}
\end{equation}

In the case of haptic memory, when the data used is reduced to one-fourth, the average value of accuracy is used as the definition to prevent overfitting.
Note that this differs from short-term memory capacity\cite{jaeger2001short}, which is defined in the context of physical reservoir computing as the ability to reconstruct delayed input sequences.
\par From here, we examine the conditions that improve haptic memory.
The heat map in \textbf{Figure \ref{fig: haptic_memory}(d)} plots the values of Equation \ref{eq: memory} and shows the distribution of the haptic memory for body parameters.
Haptic memory tends to be higher under the condition of using the body flexibly (referred to as ``flex") than under the condition of using the body rigidly (referred to as ``rigid").
To clarify the specific points in time and body parts where separability in behavior occurs, \textbf{Figure \ref{fig: haptic_memory}(e)} calculates the silhouette coefficient (see Section \ref{sec: Silhouette coefficient}) for each time series of joint angles and displays it as a color map.
Under the rigid condition, a clear separation occurs at the moment of collision, but this separation disappears quickly.
Under the flex condition, the behavior of each joint becomes unstable in response to the impact force, so the separation at the moment of the collision is not clear.
Separation occurs between the time the head moves away and the subsequent collision.
This is due to insufficient damping under flex conditions.
In addition, this long-term separation of the trajectories is more significant in the joints closer to the head under the flex condition.
This can be because the head, being the free end of the serial link system, has a greater tendency for separation.
\par To confirm the effect of propagating information to various parts of the body, we used the time series of each joint individually for learning and observed the distribution of learning performance across the body.
Each cell in Figure \ref{fig: haptic_memory}(a) uses $(x_{c2}, x_{c3}, \dots, x_{c18})^T$ as training data $\mathbf{x}_{train}([0, i\Delta t])$, while for the calculation of the accuracy distribution, $x{c2}^T, x_{c3}^T, \dots, x_{c18}^T$ are used.
\textbf{Figure \ref{fig: haptic_memory}(f)} shows the accuracy distribution in the body.
Under the flex condition, the information cannot propagate because of the body's flexibility, leading to lower learning performance on the base side.
However, on the head side, it generally surpasses the performance under the rigid condition.
Additionally, the performance when learning all joint time series simultaneously, i.e., the cell value (dashed line) under the flex condition in the heat map, exceeds that when using each joint angle time series individually (solid line).
This is likely because using angle time series from different body parts for learning simultaneously averages the noise and avoids multicollinearity.
Overall, we can conclude that haptic memory refers to the ability to propagate information related to objects to a flexible body and retain that information as behavior.
We also consider that lower body viscoelasticity (especially viscosity) and the body structure of a passive serial link with a free end enable this capability.

\subsection{Introducing the Morphological Structure}
\label{sec: Introducing the Morphological Structure}
\begin{figure}
\centering
  \includegraphics[width=0.87\linewidth]{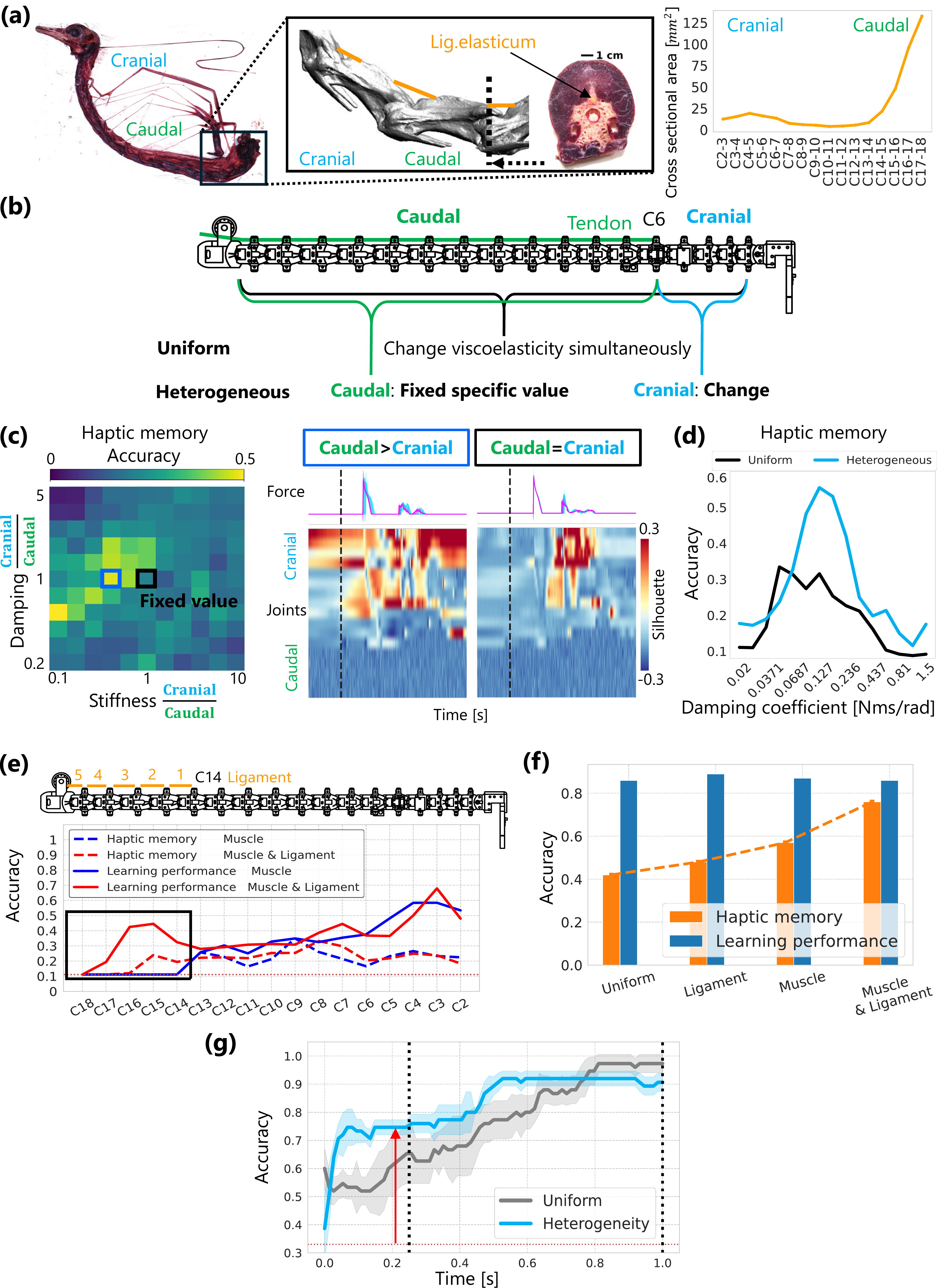}
  \caption{
Introducing the morphological structure to enhance learning performance:
(a) morphological structure in the ostrich neck, which causes the heterogeneity of bodily viscoelasticity
(b) implementation of the morphological structure in the robotic system
(c) improvement of haptic memory through the introduction of heterogeneity in body viscoelasticity
(d) relationship between damping coefficient and haptic memory.
In this figure, unlike in (f), a smoothing filter was applied to clarify the changes in body viscosity values at the maximum. Refer to the supporting information for the raw figure.
(e) effect of elastic elements for weight compensation
(f) comparison of learning performance with and without the morphological structure
(g) verification in actual robot experiments
  }
  \label{fig: improvement_by_morphology}
\end{figure}

The previous chapter demonstrated that flexible bodies inherently possess haptic memory and separability.
However, under low body viscosity conditions, learning performance tends to decrease because ESP does not hold, while haptic memory tends to increase.
This section aims to balance haptic memory and learning performance by incorporating a structure inspired by the morphology of an ostrich's neck.
As mentioned in the Introduction, tactile information is transmitted through the body's connective tissues (such as muscles and tendons).
Previous research has shown that the heterogeneous viscoelastic properties within the body contribute to the flow or guidance of information to specific parts of the body\cite{turvey2014medium}.
Therefore, we first focus on the muscle-tendon arrangement in the ostrich's neck as a factor that leads to the heterogeneity of viscoelastic properties within the body.
As shown in \textbf{Figure \ref{fig: improvement_by_morphology}(a)}, the muscles of the bird neck are divided into several groups, and we assume that the two muscle groups on the dorsal side (back side) primarily contribute to pecking movement.
Therefore, as shown in \textbf{Figure \ref{fig: improvement_by_morphology}(b)}, based on the attachment sites of these muscles, we divided the area from the beak to the C5 joint as the cranial part and from the C6 joint to the C18 joint as the caudal part.
To focus on the relative magnitudes of viscoelasticity, \textbf{Figure \ref{fig: improvement_by_morphology}(c)} presents a heat map in which the viscoelasticity of the caudal part is fixed, while the value of the cranial part varies.
Note that the vertical and horizontal axes of the heatmap represent the ratio of the caudal part values to the fixed value of the caudal part.
This heatmap shows that when the elasticity of the cranial part is smaller than that of the caudal part, the maximum value of haptic memory is achieved.
The plot on the right side of Figure \ref{fig: improvement_by_morphology}(c) is a time series of joint angles, similar to Figure \ref{fig: haptic_memory}(b).
These plots show that when the cranial part has relatively low elasticity, the separation of fluctuation trajectories becomes more distinct.
This suggests that the information is guided to the more flexible cranial region, which improves haptic memory.
\textbf{Figure \ref{fig: improvement_by_morphology}(d)} shows the relationship between body viscosity and haptic memory at the maximum value of the heat map for cases of uniform and heterogeneous body viscoelasticity.
This plot demonstrates that, under heterogeneous conditions, the maximum value of haptic memory is higher and occurs at a higher viscosity.
In this way, the heterogeneity of the body allows for higher haptic memory to be exhibited under conditions where learning performance is enhanced.
\par Next, we focus on ligaments as the elements that contribute to the heterogeneity of viscoelastic properties.
Figure \ref{fig: haptic_memory}(c) indicates that the caudal part does not significantly contribute to learning.
When the body is flexible, the force is not transmitted to the base.
Then, compensating for gravitational forces could help propagate pecking reaction forces throughout the body, potentially enhancing overall performance.
As shown in Figure \ref{fig: improvement_by_morphology}(a), a bird’s neck has ligaments known as the \textit{Ligamentum elasticum interlaminare} (Lig. elasticum), which connect adjacent joints and compensate for gravitational forces\cite{dzemski2007flexibility}.
The cross-sectional area of the Lig. elasticum increases towards the base of the neck.
To replicate this feature, we incorporate a viscoelastic element for gravitational compensation.
The supporting information presents the arrangement of ligaments in the robot and the parameters of each ligament.
The ligament differs significantly from simply changing joint stiffness in that it continuously applies force in the direction that compensates for gravity and has a steep viscoelastic distribution toward the base.
\textbf{Figure \ref{fig: improvement_by_morphology}(e)} compares the distribution of learning performance within the body under conditions with and without the ligament.
The black dashed line in the figure shows that with the ligament, gravitational compensation propagates the reaction forces to the base, enabling learning even in the caudal part.
In this way, the compensation for gravity by the ligaments increases the number of nodes in the physical reservoir, leading to improved overall learning performance.
\textbf{Figure \ref{fig: improvement_by_morphology}(f)} compares learning performance under four conditions: uniform (without morphological structures), with the ligament, with the muscle alone and with both the muscle and the ligament.
This graph shows that the morphological structure improves haptic memory while maintaining learning performance.
\par \textbf{Figure \ref{fig: improvement_by_morphology}(g)} is the validation of these findings using an actual robot.
To reproduce the heterogeneity of viscoelastic properties by muscle, Ecoflex 00-30 silicone cartilage (softer material) was used for the cranial part, while Dragon Skin 20 (harder material) was used for the caudal part.
The ligaments were implemented using metal springs.
For safety reasons, ligaments were used in both cases.
Figure \ref{fig: improvement_by_morphology}(g) shows that approximately 90\% of the learning performance was achieved in both cases.
Additionally, as indicated by the red arrow, under conditions of heterogeneity, high accuracy was observed even before the collision, demonstrating an improvement in haptic memory.
Overall, by incorporating heterogeneity in body viscoelasticity based on morphological structure, both haptic memory and learning performance can be improved.

\subsection{Real-time Inference Utilizing Musculoskeletal Dynamics}
\label{sec: Actual robot experiments}

\begin{figure}
\centering
  \includegraphics[width=\linewidth]{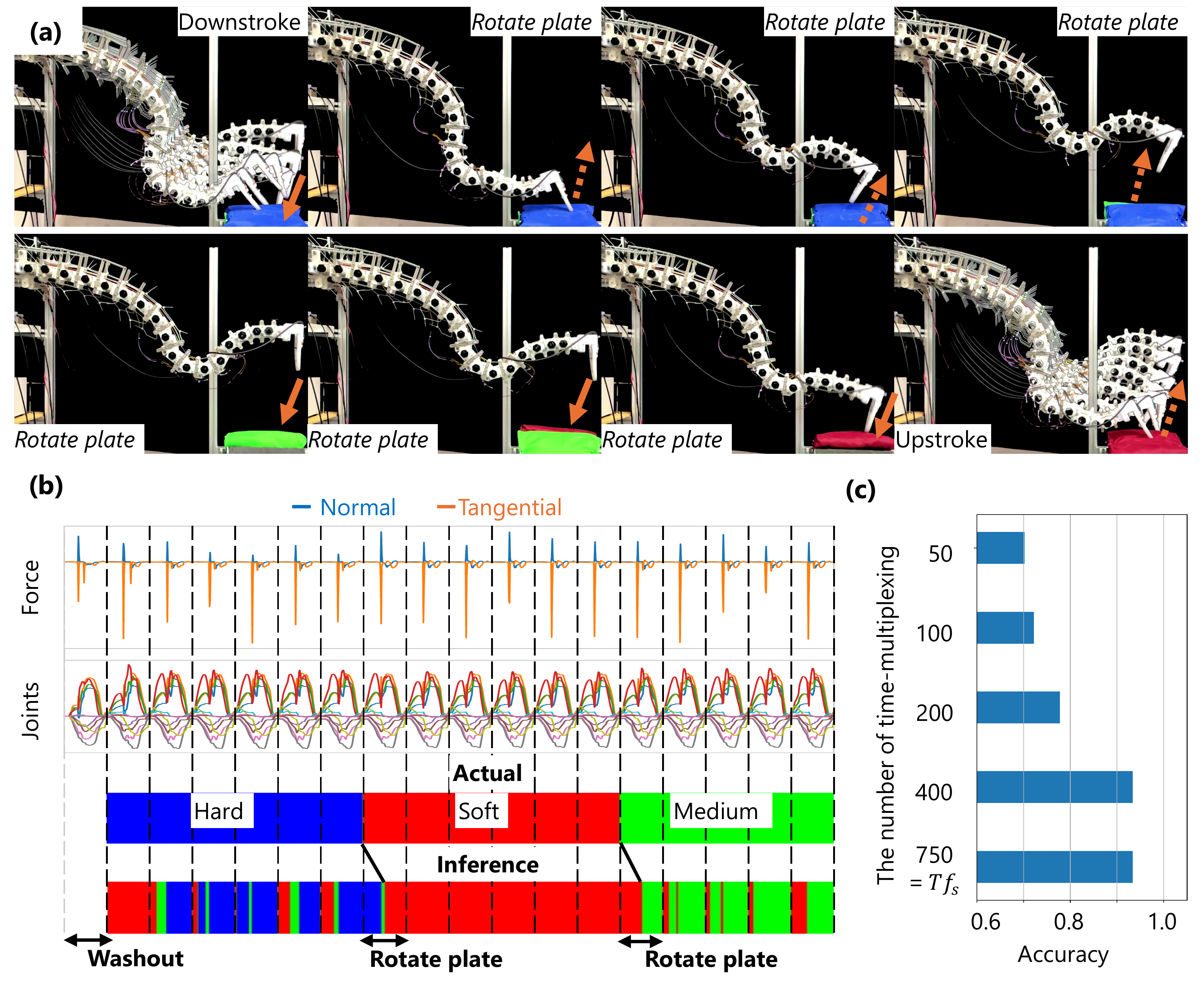}
  \caption{
Real-time inference via the dynamics of RobOstrich:
(a) pecking behavior of the robot during the real-time inference
(b) time chart of actual value and inference value
(c) evaluation of the haptic memory inherent in the body dynamics
  }
  \label{fig: Actual robot experiments}
\end{figure}

This chapter presents rapid learning using musculoskeletal dynamics and real-time inference.
The learning and inference processes follow the method described in \ref{sec: Accuracy Curve}.
At each time step, the class of the plate is inferred by sequentially applying Equation (\ref{eq: from_zero_to_it}) to the current sensor values.
\textbf{Figure \ref{fig: Actual robot experiments}(a)} illustrate RobOstrich movements during real-time inference.
The supporting information video 4 demonstrates the entire process of data acquisition, learning, and real-time inference.
The learning process took approximately 25 seconds, after which real-time inference of the object became possible.
\textbf{Figure \ref{fig: Actual robot experiments}(b)} shows the results of the real-time inference.
The upper two plots display the time series data from the force and joint angle sensors during inference, while the two color bars at the bottom compare the inferred softness of the plate (inference value) using these time series with the actual softness of the plate (actual value).
The haptic memory in the body ensures that the values of the actual value and the inference value remain consistent, even at times when the beak is separated from the plate.
As a result, during transitions in the actual value, the inference value remains the same as the previous actual value until the beak collides with the plate.
This characteristic is similar to a D flip-flop, with the reaction force acting as a clock signal and the actual value serving as the D-input.
This characteristic provides a buffering effect in the processing of impact reactions, reducing the computational load.
\par When using Equation (\ref{eq: from_zero_to_it}) for real-time inference, the $Tf_s=750$ points of the data are stored within the computer.
Using haptic memory within the body can reduce the number of data points that need to be stored on the computer.
Therefore, learning and estimation are performed using only the latest information.
The training data used is $(\mathbf{x}_{train}([(i-1)\Delta t', i\Delta t']), y_{train}(i))$, and $\hat y_j(i)$ is calculated in real time as follows:

\begin{equation}
\label{eq: realtime_estimation}
\hat y_j(i) = \sigma_i(\mathbf{x}_{test}([(i-1)\Delta t', i\Delta t']))
\end{equation}

A small $\Delta t'$ means that fewer data points need to be stored on the computer, and when $\Delta t' = T_sf$, it means that all the data for one cycle are stored on the computer.
\textbf{Figure \ref{fig: Actual robot experiments}(c)} show the accuracy with respect to $\Delta t'$.
This accuracy refers to the proportion of times that "Actual" and "Inference" match on the color bar in Figure 8(b).
Time periods corresponding to washout and rotate plate are not taken into account.
Up to $\Delta t = 50$, that is, even by reducing the amount of information stored on the computer to one-fifteenth, a 70\% accuracy can be maintained.
These results demonstrate that physical reservoir computing not only enables rapid training, but also leverages the body's inherent haptic memory to significantly reduce the computational load during inference.

\section{Conclusion}
\label{sec: conclusion}
This study proposes a method utilizing physical reservoir computing to leverage musculoskeletal dynamics to address two challenges in haptic perception for soft robots: the fragility of their bodies and the computational cost of learning self-deformation.
We conducted a comprehensive physical simulation to identify the conditions that enable high learning performance among the diverse behaviors a flexible body can exhibit.
The simulation results revealed that, for high learning performance, the behavior must exhibit an asymptotic stability property known as the echo state property.
We demonstrated that a body with appropriate viscoelasticity mitigates impact forces during collisions and stably amplifies input differences, thus improving learning performance.
Traditionally, soft materials located at the boundary between the body and the environment (i.e., skin) have been considered to play a role in mitigating impact forces\cite{shimoga1996soft} and enhancing external force information\cite{Cutkosky1993}.
Our results indicate that a flexible musculoskeletal system can substitute for these functions, providing a potential solution to the fragility inherent in robots primarily composed of soft materials.
\par Next, we demonstrated that under conditions of rapid input and flexible body parameters, information from the previous cycle of periodic movements can be retained as behavior.
We quantified this memory using physical reservoir computing and showed that incorporating the heterogeneity of body viscoelasticity, inspired by the arrangement of muscles and ligaments in an ostrich’s neck, enhances this memory.
The evaluation of computational capabilities of biological anatomical structures using reservoir computing is gaining attention as a promising approach for mutual inspiration between biology and computational science\cite{Luís2019,Valero2007}.
The actual ostrich neck consists of about 200 muscles and a complex system of force transmission through tissues with varying material properties\cite{Zweers1994,cobley2013inter}, such as fascia and tendons, suggesting the potential for even more diverse computational capabilities.
Our results represent an initial step in exploring this potential.
Based on these findings, we propose a haptic perception system that enables rapid learning and real-time inference by utilizing musculoskeletal dynamics.
This task is analogous to how bipedal animals recognize ground conditions during running and hopping\cite{shimoga1996soft,moritz2004passive} .
In such cases, animals must rapidly adapt to changes in surface properties under the constraints of sensorimotor delays (e.g., nerve transmission delays and muscle electromechanical coupling) to prevent falling\cite{Monica2018}.
The information from the previous cycle, retained within the body, is considered useful for generating predictive feedforward commands that enable adaptive behavior.
In this study, the input and body parameters for haptic perception were predefined manually.
Therefore, in future work, we propose to build a closed-loop system and develop mechanisms and algorithms for parameter adjustment.
Recently, Augmented Direct Feedback Alignment, which replaces state updates based on the gradient of nonlinear mappings with random matrices, has been proposed\cite{nakajima2022physical}.
We are considering using this algorithm to adjust the body parameters.
\par It is important that interactions between the biological system and the environment can lead to the emergence of functions beyond the original design intent.
When animals adapt to new environments, they do not necessarily acquire new body parts.
Due to phylogenetic, developmental, and genetic constraints, they derive new functions from existing structures to gain new functionalities\cite{fukuhara2022comparative}.
For robots, the ability to adapt to diverse environments using a finite body is also crucial.
Recently, mechanical implementations of combinational and sequential logic circuits, the fundamental units of information processing, have been proposed to achieve embodied intelligence\cite{yan2023origami,Daniel2019}.
In contrast, this study considers a flexible body originally designed as a manipulator as a physical reservoir, achieving sensing and memory functions by learning the readout without dedicated modules.
Expanding physical reservoir computing for biological systems could further enable the perception of diverse self- and environment-related information, leading to emergent functions.

\section{Experimental Section}

\subsection{Experimental Setup in Physical Simulator}
\label{sec: Experimental Setup in Physical Simulator}
The setup for the computational experiments is shown in the supporting information.
As mentioned above, haptic perception based on musculoskeletal deformation can be interpreted as the acquisition of time-invariant values that arise through dynamic interaction with the environment\cite{turvey2014medium}.
Therefore, this haptic perception is emulated as a task to learn the dynamics of a flexible body and to determine the virtual damping coefficient $D$ of the plate as shown in Figure \ref{fig: Physical_reservoir}(c).
In this case, dynamics exists not only in the motion of the plate but also in the contact between the beak and the plate.
Specifically, the reaction force is generated in response to the amount and speed of the beak that penetrates the board (constraint violation).
Then, the dynamics of the plate is estimated through the dynamics of this contact.
To emulate this, we used a physical simulator, Mujoco (multi-joint dynamics with contact)\cite{todorov2012mujoco}.
It is specialized for computing of multi-joint structures interacting with the environment and can replicate the slack in the tendon, allowing for the reproduction of the swinging down movement of the neck in accordance with the natural dynamics.
This experimental environment is based on the RobOstrich 3D CAD model that we developed previously\cite{nakano2023robostrich}.
The calculations of the interactions are valid only for the joint ranges of each joint and for the contact between the beak and the plate, ensuring that the plate does not collide with anything other than the beak, as shown in Figure \ref{fig: Physical_reservoir}(c).
This simulation environment can set viscosity (damping coefficient) and elasticity (stiffness) independently for each joint, which determines the unique dynamics of the neck when the tendon is slackened.
In the simulation setup, \( f_s = 700 \, \text{Hz} \) and \( \Delta t = 10 \) were used, taking into account the consistency with the force sensors in the actual robot and the mechanical receptors in the biological system.
Additionally, to prevent overfitting, the sensor data was quantized to 12 bits.

\subsection{Experimental Setup in an Actual Robot}

\label{sec: Experimental Setup in Actual Robot} 
For physical validation, we used the actual RobOstrich that we previously developed\cite{nakano2023robostrich}. Magnetic encoders (AS5600, ams OSRAM) are installed in each joint, and a force sensor (MMS101, Minebeamitsumi) is installed at the tip of the beak. 
In biological systems, the method of sensing joint angles has long been a topic of controversy\cite{gibson1966senses, mccloskey1978kinesthetic}. 
A recent consensus suggests that information from muscle receptors plays a more dominant role than that from joint receptors\cite{Proske2012}. 
On the other hand, in tendon-driven musculoskeletal robots, joint angle sensors can be installed at each joint, providing a rational approach that allows easy detection of self-position while maintaining the flexibility of the body\cite{kawaharazuka2019}. 
The tendon is made of high-strength polyethylene fiber wire (Dyneema ULTRA2 No.15) and is arranged to pass through each joint using ceramic bushes. 
The endpoint of the tendon is connected to the C6 joint (head-side joint, see Figure \ref{fig: Physical_reservoir}(b)), while the other endpoint is pulled through a pulley by a low reduction ratio electric motor (AK80-6, T-motor). 
Adjustments must be made in advance before the experiments in all cases. 
The methods for adjusting the viscoelastic properties of each joint are shown in Figure \ref{fig: Physical_reservoir}(d) and is described as follows. 
we use silicon cartilage in each joint to adjust the viscoelasticity. 
The viscoelastic properties can be changed by varying the material of the cartilage. 
In this paper, Dragon Skin 20 and Ecoflex 00-30 were used. 
Representative mechanical properties are shown in the supporting information. 
Additionally, we adopted a design in which springs and dampers can be individually attached to each joint to separately evaluate the effects of the body's viscosity and elasticity. 
Friction-type brakes (MK08, Tsubakimoto Chain Co.) installed in each joint can generate friction torque. 
The magnitude of this frictional torque can be set according to the amount of rotation of a knob. 
Although friction remains constant with respect to the velocity of the joints, the viscosity is proportional to the velocity of the joints. 
However, both have the same effect of dissipating the kinetic energy, enabling qualitative reproduction of the behavior. 
As shown in Figure \ref{fig: Physical_reservoir}(c) and \ref{fig: Physical_reservoir}(d), the pecked object consists of soft polyurethane material and hard silicone rubber sheets, all wrapped in nylon fabric.
The arrangement of the rubber sheets varies with the color of the nylon (blue for 5 mm, green for 10 mm, and red for 20 mm), resulting in progressively harder objects in the RGB color order. 
Note that it cannot classify them solely based on the robot's posture at the moment of contact, as the stress-strain characteristics should be the same when a static uniform load is applied to the object in this setup. 
The three types of object are mounted on a vertical electric turntable, allowing for rapid switching of objects without hindering the movement of the RobOstrich.
These systems are controlled by a general-use mini PC (HX90, MINISFORUM). 
In actual robot experiments, \( f_s = 750 \, \text{Hz} \) and \( \Delta t = 10 \) were used.

\subsection{Softmax Classifier}
\label{sec: Softmax Classifiers}
This section describes in detail the softmax classifier used as the readout for the physical reservoir computing.
For the $K$ classes, let the linear softmax score for each class k be denoted as $s_k(\mathbf{x})$.
$s_k(\mathbf{x})$ can be written as follows:

\begin{equation}
\label{eq: softmax score}
s_k(\mathbf{x}) = \mathbf{x}^T\theta^{(k)}
\end{equation}

Here, $\mathbf{x}$ represents the sensor time series data and $\theta$ is the parameter vector.
By applying the softmax function to $s_k(\mathbf{x})$, the probability $\hat p_k$ that $\mathbf{x}$ belongs to class $k$ can be calculated.

\begin{equation}
\label{eq: softmax}
\hat p_k = \sigma(s_k(\mathbf{x}))_k=\frac{\text{exp}(s_k(\mathbf{x}))}{\sum_{j=1}^{K} \text{exp}(s_j(\mathbf{x}))}
\end{equation}

In classification, the class with the highest probability is estimated as follows:
\begin{equation}
\label{eq: probability estimation}
\hat{y} = \arg\max_{k} \sigma(s_k(\mathbf{x}))
\end{equation}

In training, the parameters $\theta$ that optimize the loss function known as cross-entropy, as shown below, are determined using gradient descent.

\begin{equation}
\label{eq: learning}
J(\theta)=-\frac{1}{M} \sum_{m=1}^{M} \sum_{k=1}^{K} y_k^{(m)} \log(\hat p_k^{(m)})
\end{equation}

As shown in Equation \ref{eq: softmax} to Equation \ref{eq: learning}, softmax regression is one of the simplest classification methods.
Therefore, when considering the dynamics of a complex system as a physical reservoir, it is suitable for evaluating its computational capacity by attaching a readout.

\subsection{Silhouette Coefficient}
\label{sec: Silhouette coefficient}

Silhouette analysis is a method used to evaluate the performance of clustering algorithms\cite{rousseeuw1987silhouettes}. 
The silhouette coefficient is a measure of how similar an object is to its own cluster (cohesion) compared to other clusters (separation).
For each data sample $x(i)$, the silhouette coefficient is calculated using the following steps:

Cohesion within a cluster is measured by calculating the average distance from $x(i)$ to all other points in the same cluster $C_{in}$:

\begin{equation}
    a(i) = \frac{1}{|C_{in}| - 1} \sum_{x(j) \in C_{in}} \| x(i) - x(j) \|
\end{equation}

Separation from other groups is measured by calculating the average distance from $x(i)$ to the points in the nearest different cluster $C_{near}$:

\begin{equation}
    b(i) = \frac{1}{|C_{near}|} \sum_{x(j) \in C_{near}} \| x(i) - x(j) \|
\end{equation}

Here, $|C|$ represents the number of points in cluster $C$.
Using the larger of $a(i)$ and $b(i)$, the silhouette coefficient is defined as

\begin{equation}
    s(i) = \frac{b(i) - a(i)}{\max(a(i), b(i))}
\end{equation}

By definition, the silhouette coefficient ranges from $[-1,1]$.
Values close to 1 indicate better clustering performance.
Values near 0 indicate overlapping clusters. 
Negative values generally indicate that a sample has been assigned to the wrong cluster.
Since the silhouette coefficient may not be calculated correctly, Gaussian noise is added for the calculation in this study.

\medskip

\medskip
\section*{Acknowledgements} 

This work is supported by JSPS KAKENHI Grant Numbers 24KJ0599 and 21KK0182 and 23K18472 and by JST CREST Grant Number JPMJCR2014.
K. I. was supported by JSPS KAKENHI Grant Number 22K21295.
The authors thank Dr. Megu Gunji for providing photographs used in this article.

\section*{Ethics Approval}
This paper includes anatomical illustrations of an ostrich. The specimen used was obtained from an animal processed for food purposes, not specifically for research, and therefore its use does not involve any ethical issues related to animal experimentation.

\bibliographystyle{unsrt}
\bibliography{main}

\end{document}


\maketitle

\newcommand{\secbehavior}{2.3}  
\section{Behaviors that Enable Haptic Perception}
\label{subsec: Relationship between haptic perception and behaviors}

\begin{figure}[H]
\centering
  \includegraphics[width=\linewidth]{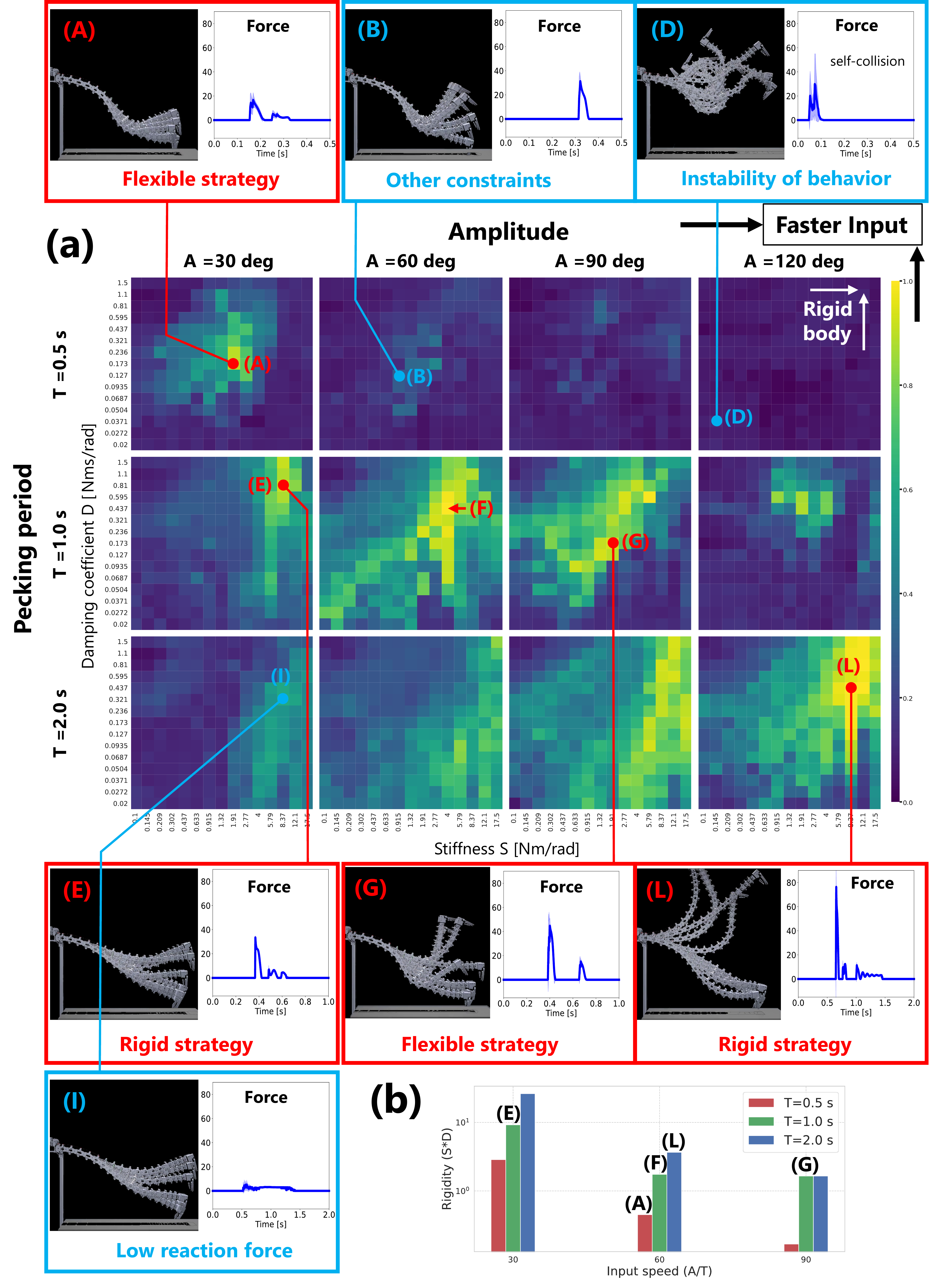}
  \caption{
   Relationship between input, physical parameter and task performance
    }

  \label{fig: heatmaps_and_behaviors}
\end{figure}

In this section, we analyze the distribution of learning performance across all combinations of input and body parameters, as explained in Section $\secbehavior$ of the main text.

\textbf{Figure \ref{fig: heatmaps_and_behaviors}(a)} illustrates the distribution of the maximum value of the accuracy curve with respect to both physical and input parameters.
Each plot consists of an inner heatmap representing 225 variations in body viscoelasticity, derived from 15 levels of elasticity ($S$) and 15 levels of viscosity ($D$).
These heatmaps are embedded within an outer matrix of 12 input configurations, based on combinations of three oscillation periods ($T$) and four amplitudes ($A$).
Within each inner heatmap, the upper-right regions correspond to stiffer body conditions (higher $S$ and $D$ values), while in the outer matrix, the upper-right cells represent faster input-driven movements (i.e., larger $A$ and shorter $T$).
Note that the actual pecking period of an ostrich is approximately 0.5 to 1.0 s.
The surrounding figures illustrate behavioral patterns and the corresponding reaction force time series for selected heatmap cells, particularly those located at local maxima.

The outer matrix reveals that learning performance degrades under both excessively fast (D) and excessively slow (I) input conditions.
In condition (D), the input speed exceeds the viscoelastic responsiveness of the body, thus preventing stable pecking.

In contrast, condition (I) generates an inadequate reaction force, leading to insufficient reaction force from the pecked object.
These findings emphasize the role of physical and informational constraints in shaping learning outcomes.
In conditions (D) and (I), the constraints that hinder learning performance are apparent.
However, in condition (B), stable pecking is achieved, and the maximum reaction force is similar to that observed in successful learning conditions, such as (A), (E), (G), and (H). Despite this, learning performance remains low.
Even when the potential for learning is not immediately apparent from behavior, physical and informational constraints may still hinder effective learning.

\par We next examine the conditions associated with high learning performance
Due to physical and informational constraints, the regions of high accuracy in the inner heatmaps tend to align diagonally across the outer matrix.
Conditions such as (A), (E), (F), (G), and (L) correspond to local maxima in these heatmaps and demonstrate effective task performance.
The physical parameters at these maxima vary according to the specific input settings, reflecting a strong coupling between body dynamics and control inputs.

To investigate this relationship more formally, we plotted input speed (defined as amplitude divided by period, $A / T$) against body stiffness (quantified as the product of elasticity and viscosity, $S × D$) at the points of maximal task performance.
The resulting graph, shown in \textbf{Figure \ref{fig: heatmaps_and_behaviors}(b)}, displays a generally negative slope.
This trend reveals a trade-off between input and body parameters in achieving optimal performance in haptic perception.
In other words, effective strategies include either driving a flexible body at high speed (e.g., A, G) or operating a rigid body at lower speed (e.g., E, L).
Section $\secbehavior$ of the main text provides a comprehensive summary of the interdependencies between input characteristics, body parameters, and learning performance.

\newpage

\newcommand{\ESPindex}{2.4}  
\section{ESP index}

\begin{figure}[H]
\centering
  \includegraphics[width=\linewidth]{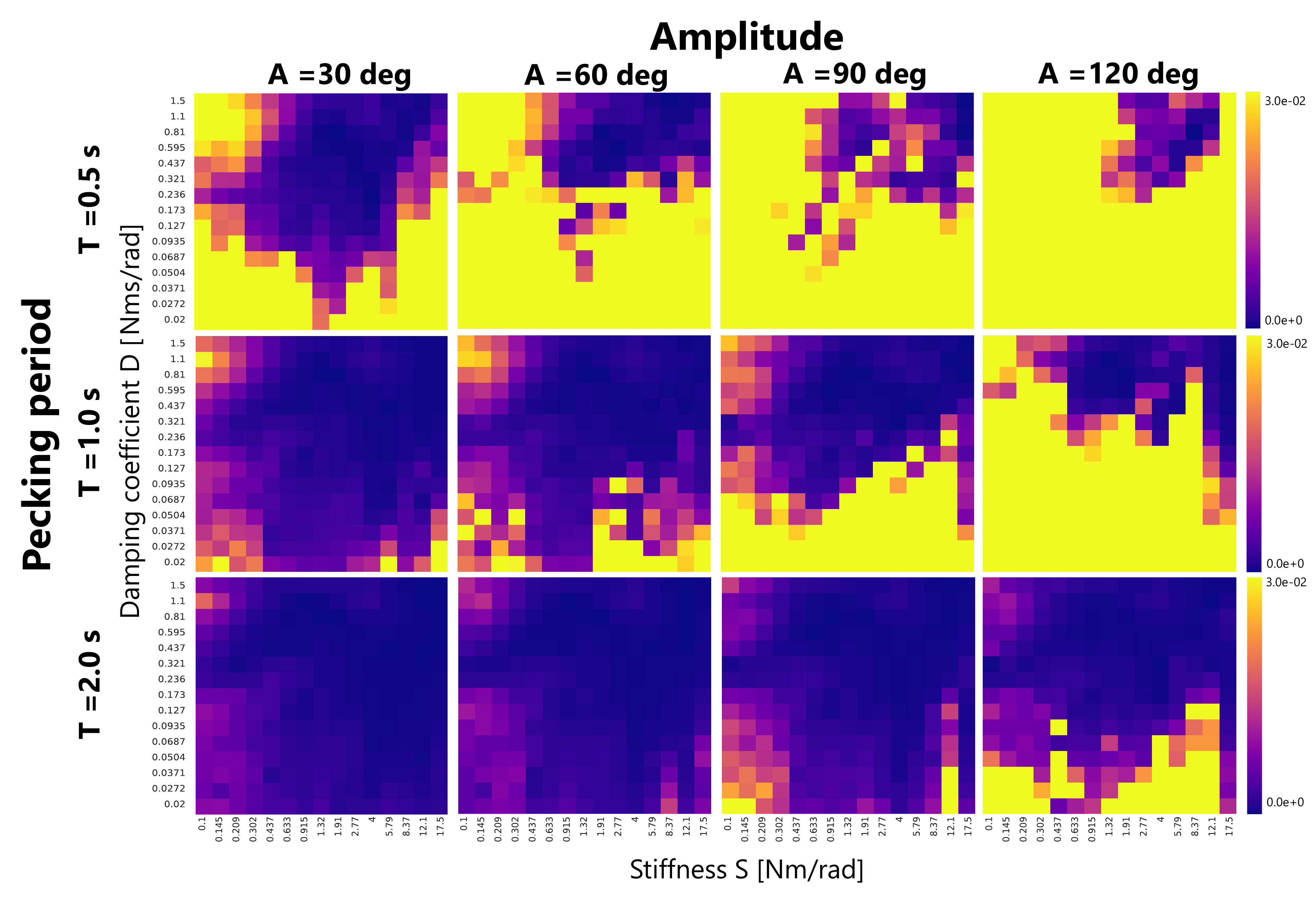}
  \caption{
   Distribution of Echo State Property Index
    }

  \label{fig: ESP_heatmaps}
\end{figure}

In this paper, we focused on a specific set of input parameters ($A = 90^\circ$ and $T = 1.0$ s).
\textbf{Figure \ref{fig: ESP_heatmaps}} shows the distribution of the ESP index for all input parameters.
Compared with Figure \ref{fig: heatmaps_and_behaviors}, the following points are observed: 

\begin{itemize} 
\item There is a clear relationship between learning performance and the echo state property (ESP); performance improves under conditions where the ESP holds. 
\item The ESP breaks down when there is a significant imbalance between the body’s viscosity and elasticity. 
\end{itemize}

Since the ESP is associated with the stability of behavior, it is reasonable that learning performance improves in regions where the ESP is valid.
Interestingly, even within ESP-valid regions, performance tends to be highest near the boundary where the ESP begins to break down.
This trend is especially evident in areas with relatively low body viscoelasticity ($S < 2.77$ [Nm/rad]).
This indicates that enhancing learning performance requires the body to exhibit a certain level of sensitivity to external forces without becoming unstable.

\newpage

\section{Separability}

\begin{figure}[H]
\centering
  \includegraphics[width=0.84\linewidth]{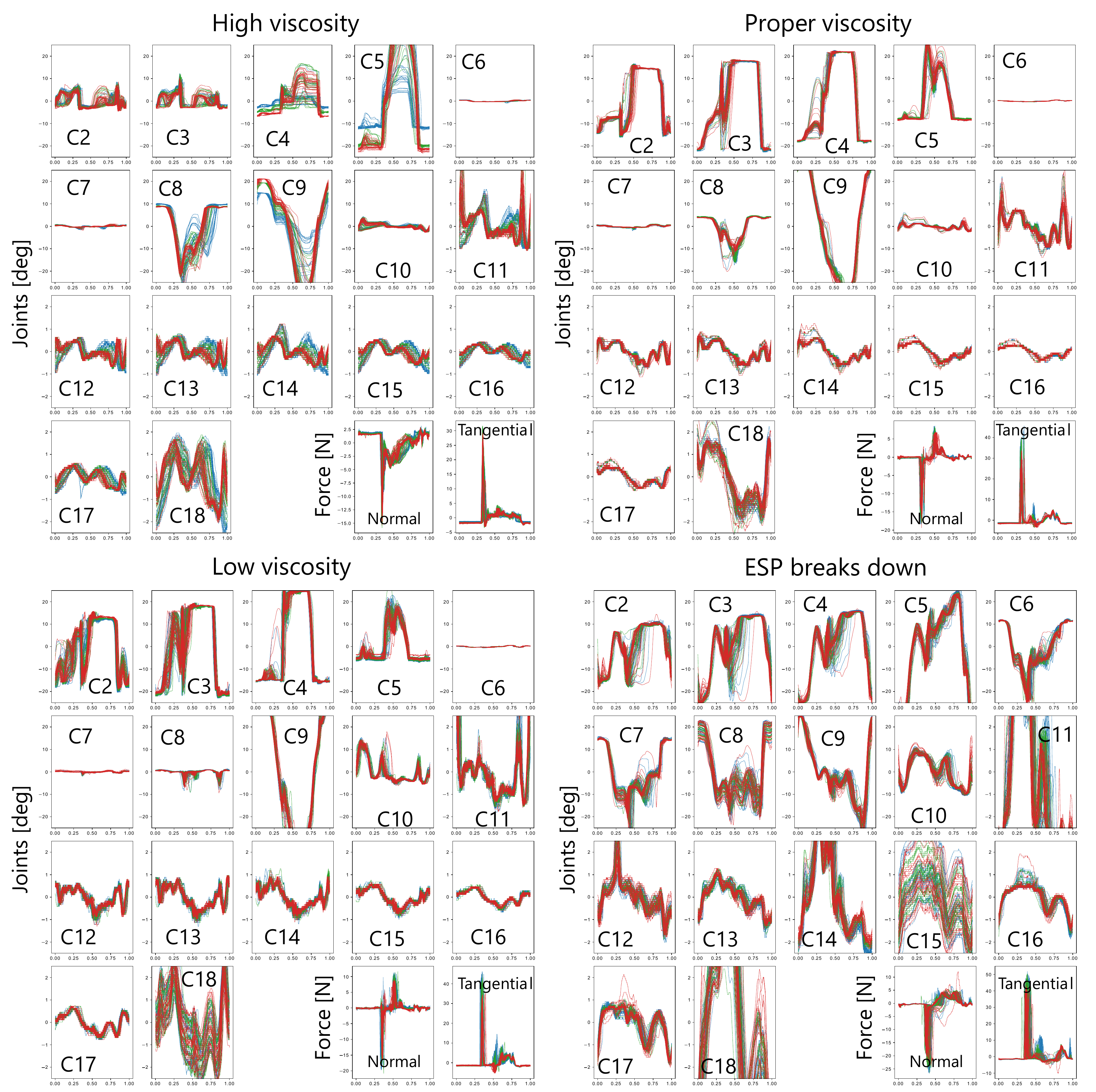}
  \caption{
   The dynamics (joint angles, reaction forces) when changing body viscosity
    }

  \label{fig: separability}
\end{figure}

In the left panel of Figure 5 (d) in main text, we compare the accuracy curves across varying input speeds and body viscoelasticity.
Figure \ref{fig: separability} in this document shows the corresponding joint angle and reaction force time series. 
Each color indicates a different target object that was pecked.
In the joint angle plots, the top-left panels correspond to the head and the lower panels correspond to the base.
The bottom-right panels display the reaction force time series.
These plots represent conditions with High, Proper, and Low body viscoelasticity, as well as a case with increased input speed (leading to ESP breakdown).
In the high viscoelasticity condition, the force separation is unclear, which results in lower learning performance.
With proper viscoelasticity, the trajectory remains more stable and a clear post-peak force separation emerges, yielding the best performance.
In the case of low viscoelasticity, instability in the joint angle trajectories leads to increased deviation in the force signals.
Finally, a higher input speed increases the deviation in both signals, causing the lowest performance.

\newpage

\section{Haptic Memory}

\begin{figure}[H]
\centering
  \includegraphics[width=\linewidth]{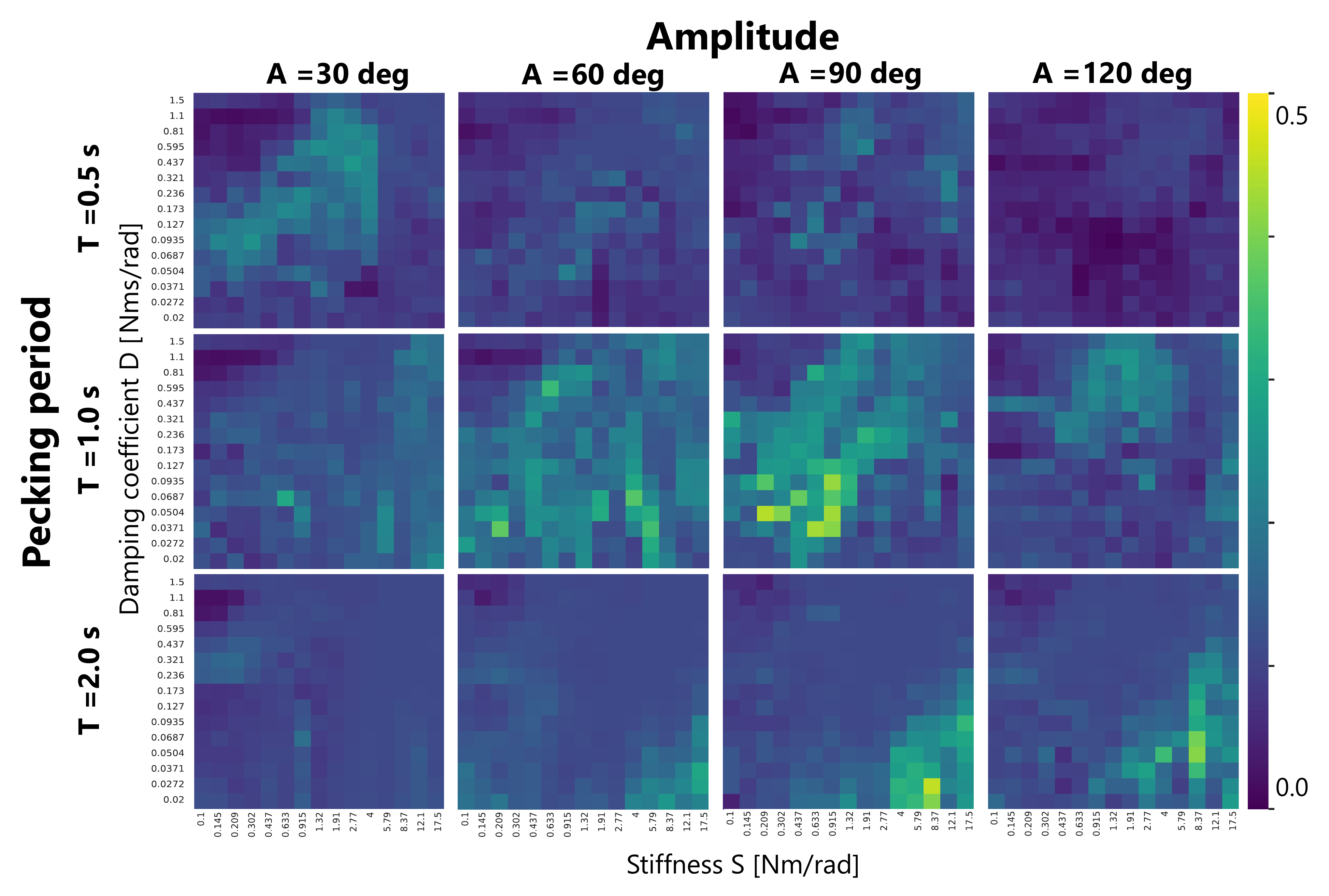}
  \caption{
   Distribution of haptic memory
    }

  \label{fig: heatmap_memory}
\end{figure}

As mentioned in the main text, the definition of haptic memory is similar to that of learning performance and is defined as the accuracy when learning data are limited to the time series before the collision.
\textbf{Figure \ref{fig: heatmap_memory}} shows the distribution of the input and body parameters, as with the learning performance.
Note that the upper limit of the heatmap is 0.5.
As can be seen from the comparison between Figure \ref{fig: heatmaps_and_behaviors} and Figure \ref{fig: heatmap_memory}, under input conditions with low learning performance, haptic memory also tends to be low.
However, haptic memory does not fully correspond to learning performance, as it is observed that haptic memory takes high values under conditions with low viscosity.
Therefore, in the subsequent discussion, we will address the heterogeneity of body viscoelasticity, inspired by the structure of the ostrich's neck.

\newpage

\section{Introducing the morphological structure}

\begin{figure}[H]
\centering
  \includegraphics[width=0.8\linewidth]{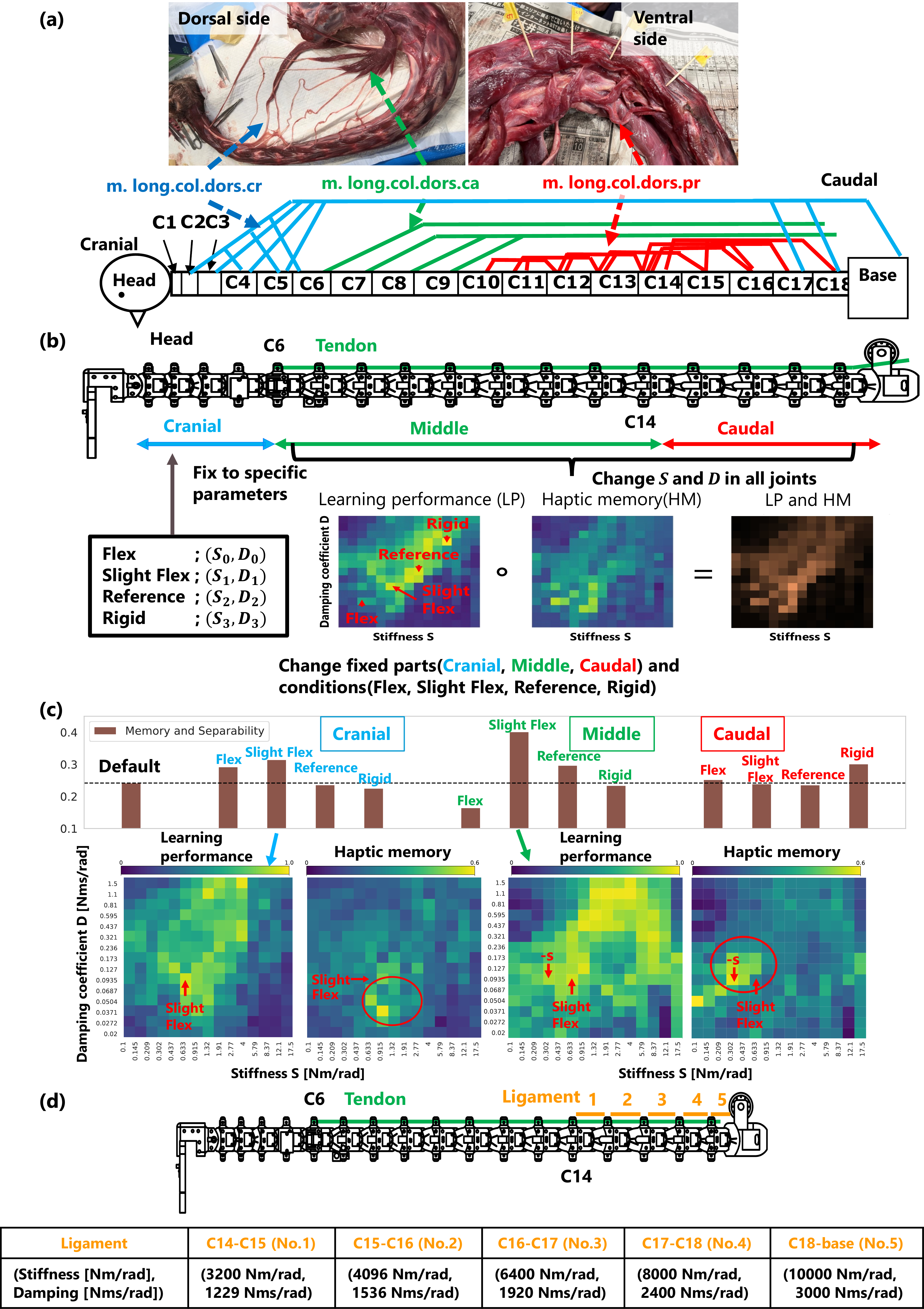}
  \caption{
   Introducing morphological structure
   (a) Muscle arrangement of an ostrich
   (b) Exploration of internal viscoelasticity patterns of the body
   (c) Fixed positions and fixed values that achieve both learning performance and haptic memory
   (d) Parameters in each ligaments
    }

  \label{fig: Introducing morphological structure}
\end{figure}

\newpage

In the case of a multi-degree-of-freedom structure, the combinations of viscoelasticity at each joint are numerous, making it difficult to determine the optimal distribution. 
Therefore, we focus on the muscle-tendon configuration of an ostrich's neck.
As described in the main text, we divide the neck into cranial and caudal regions. 
In main text, we fixed the parameters on the caudal side and varied those on the cranial side to investigate the heterogeneity of viscoelasticity in the body.
However, as shown in \textbf{Figure \ref{fig: Introducing morphological structure} (a)}, 
the avian neck can actually be categorized into three distinct regions based on muscle arrangement and joint mobility. 
This structural distinction requires a reconsideration of which region's values should be fixed.
As shown in \textbf{Figure \ref{fig: Introducing morphological structure} (b)}, 
we divided the neck into three parts according to muscle configuration:
the cranial part (from the beak to the C5 joint), 
the middle part (from the C6 to the C13 joint), 
and the caudal part (from the C14 to the C18 joint).
We defined four fixed-value conditions (F0, F1, F, F2), 
each corresponding to viscoelasticity parameters that yielded peak performance under the input setting (A = 90 deg, T = 1.0 s).
For each experiment, one region was fixed to a specific condition, while the other two regions were varied across 225 combinations of viscoelasticity parameters.
We generated heatmaps for both learning performance (LP) and haptic memory (HM), 
and then computed the element-wise product to create a combined heatmap (LP and HM), representing the joint optimization of both metrics.
A high value in the (LP and HM) heatmap indicates that both learning performance and haptic memory are achieved simultaneously.
\textbf{Figure \ref{fig: Introducing morphological structure} (c)} compares the maximum values of combined learning performance and haptic memory ((LP and HM)) under each fixed condition.
Notably, when the caudal part is fixed (indicated by the blue bars), the performance shows minimal improvement. 
This suggests that tuning the cranial and middle parts has a more significant impact.
Furthermore, whether the fixed region is the cranial part or the middle part, performance is maximized when the fixed value corresponds to the “F1: slight flex” condition.
The bottom row of heatmaps in Figure \ref{fig: Introducing morphological structure} (c) shows the learning performance and haptic memory under the corresponding conditions.  
The two heatmaps on the left correspond to the case where the cranial part is fixed at the “slight flex” condition.  
Looking at the heatmap of learning performance, we see that fixing the cranial region—where the ESP is satisfied—at a slightly flexible value results in higher learning performance over a wider range, compared to when the body viscoelasticity is uniform.  
In the haptic memory heatmap, we observe improvements in the region to the right of the fixed value (highlighted by the red circle).  
This suggests that haptic memory improves when the cranial region is relatively flexible.  
This may be due to the separation induced on the cranial side, as RobOstrich has a structure with a free end.  
The two heatmaps on the right show the case where the middle part is fixed at the “slight flex” condition.  
A similar discussion applies in this case as well.

From this discussion, we draw the following key conclusions:
\begin{itemize}
  \item \textbf{Minimal effect of caudal region:}  
  Changes in the viscoelasticity of the caudal part have little influence on learning performance and haptic memory.  
  Therefore, the neck can effectively be treated as two regions: cranial (C6 and above) and caudal (below C6).

  \item \textbf{Optimal fixed value:}  
  The best results are achieved when the fixed value is slightly more flexible than the global maximum in the heatmap.  
  Specifically, \( S = 0.633\ \mathrm{Nm/rad},\ D = 0.0935\ \mathrm{Nms/rad} \).

  \item \textbf{Design principle:}  
  To improve haptic memory, the cranial region should be designed to be more flexible than the caudal region.
\end{itemize}

These findings are summarized in the main text.

\textbf{Figure \ref{fig: Introducing morphological structure} (d)} shows the ligament parameters used in the simulator. 
Inspired by the cross-sectional areas of the ostrich ligaments, the parameter values increase geometrically toward the base of the neck.
Unlike simply adjusting joint parameters, these ligaments continuously apply force in a direction that compensates for gravity.

\newpage
\begin{figure}[H]
\centering
  \includegraphics[width=0.3\linewidth]{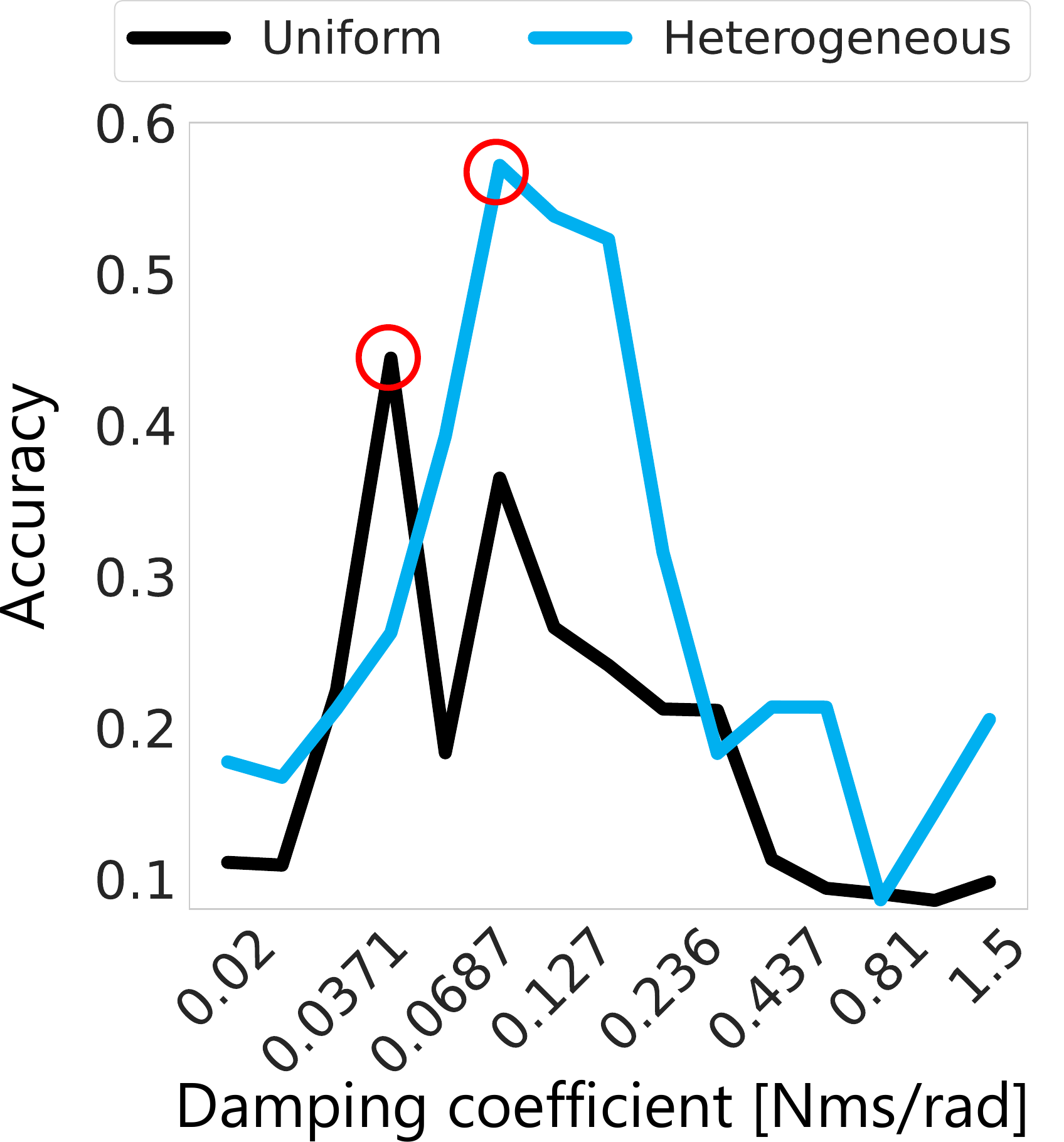}
  \caption{
    The graph in Figure 7 (d) before smoothing, which illustrates the improvement of haptic memory.
    }
  \label{fig: before smoothing was applied}
\end{figure}

\textbf{Figure~\ref{fig: before smoothing was applied}} shows the version of Figure 7(d) before smoothing was applied.
Learning performance can exhibit abrupt increases or decreases when body viscoelasticity is varied, due to qualitative changes in behavior.
In such cases, the behavior at a particular viscosity is not necessarily intermediate between its neighboring values.
Therefore, applying smoothing to this graph would not appropriately reflect the underlying dynamics.
However, to clearly demonstrate that introducing heterogeneity in body viscoelasticity increases both the maximum accuracy and the corresponding viscosity value, smoothing was applied in Figure 7(d).
This increase in viscosity helps prevent ESP breakdown while improving memory.
The red circles in Figure 7(f) of the main text indicate the corresponding maxima.

\newpage

\begin{figure}[H]
\centering
  \includegraphics[width=0.9\linewidth]{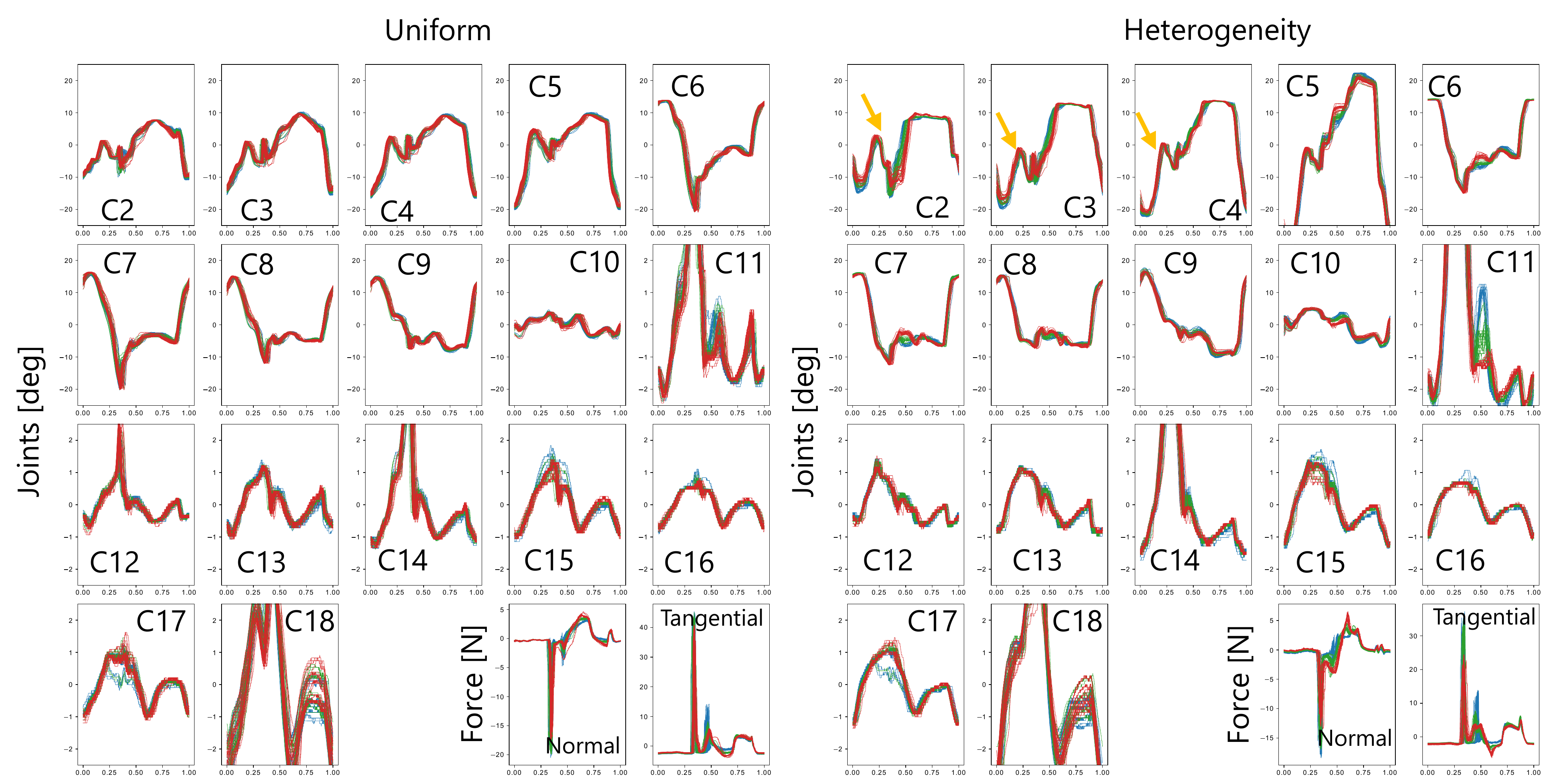}
  \caption{
    Comparison of dynamics in cases of uniform and heterogeneous body viscoelasticity
    }
  \label{fig: heterogeneity}
\end{figure}

\noindent
As discussed in the main text, haptic memory refers to the robot’s ability to separate its trajectory depending on the softness of the object it pecks at. This separation enables long-term learning. There is a trade-off between haptic memory and learning performance, particularly in relation to the viscoelasticity of the body.

The simulation results show that lowering the elasticity of the head can guide displacement, improving haptic memory without reducing learning performance. This point is also experimentally demonstrated in Figure 7 (g) of the main text, using the accuracy curve.

\textbf{Figure \ref{fig: heterogeneity}} shows the time series of joint angles from the accuracy curve in Figure 7(g). The colors represent different types of pecked objects. The left panel shows joint angles and reaction forces when all joints have silicone cartilage made from Dragonskin 10. This results in uniform viscoelasticity throughout the body. In contrast, the right panel shows the case where only the head joint’s silicone cartilage is made from Eco-flex 30. This introduces heterogeneity in the viscoelasticity of the body, making the head softer.

Looking at the joint angles, the trajectory separation is smaller in this case compared to Figure \ref{fig: separability}, where only friction brakes were used without silicone cartilage. This difference is likely because friction brakes cause a delay in generating a reaction force in response to the angular velocity of the joint.

When comparing the left and right panels, we can see that trajectory separation occurs, particularly in the head joint, before contact. This is highlighted by the orange arrows. This trajectory separation corresponds to the difference in haptic memory observed in Figure 7(g) of the main text.

\section{Additional information of physical simulator experiments}

\begin{figure}[H]
\centering
  \includegraphics[width=0.9\linewidth]{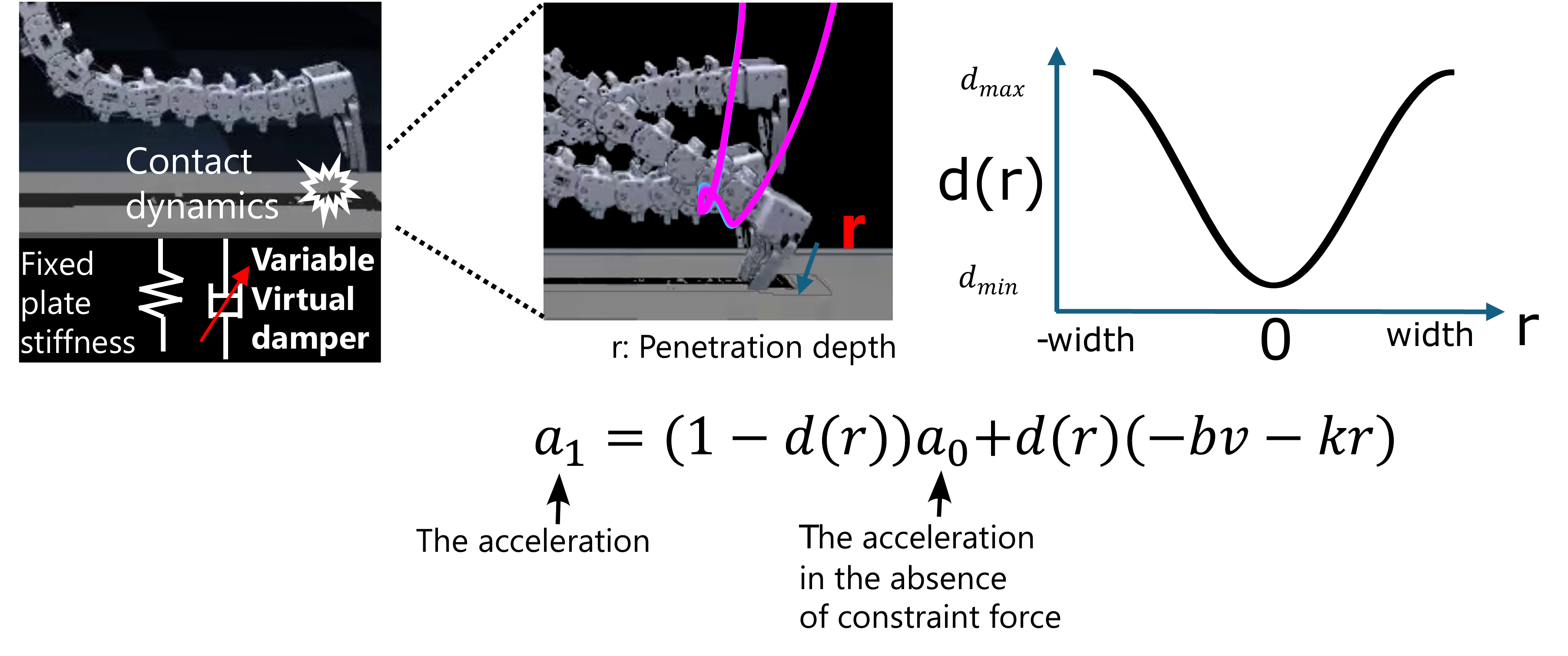}
  \caption{
Detail of contact computation
    }
  \label{fig: contact_dynamics}
\end{figure}

As described in Figure 1(c) of the main text, this task involves estimating the underlying dynamic properties (a virtual damping coefficient) through contact dynamics. 
Here, we explain how contact dynamics is computed in the simulator.

This chapter follows the MuJoCo documentation at \url{https://mujoco.readthedocs.io/en/stable/modeling.html#csolver}, accessed on April 6, 2025.

\par MuJoCo models constraints using virtual spring-damper systems, and computes constraint forces (reaction forces) to enforce them. 
The dynamics of constraint is approximately governed by:

\[
a_1 + d \cdot (b v + k r) = (1 - d) \cdot a_0
\]

Where:
\begin{itemize}
    \item \(a_1\): acceleration with constraint force,
    \item \(a_0\): acceleration without constraint force,
    \item \(v\): velocity,
    \item \(r\): constraint violation (e.g., penetration depth),
    \item \(d\): impedance (ability of the constraint to generate force),
    \item \(b\): damping,
    \item \(k\): stiffness.
\end{itemize}

The parameters that are under user control are $d,b,k$.

Impedance \(d \in (0, 1)\) controls how strongly the constraint responds. Small \(d\) implies a soft constraint; large \(d\) implies a stiff constraint. 

The impedance can be constant or vary with \(r\).
This defines a smooth sigmoid function \(d(r)\), which increases with \(|r|\), allowing for position-dependent constraint strength.
In this study, following the default settings, we set the order of the sigmoid function (power value) to 2 and the inflection point (midpoint value) to 0.5.
As shown in \textbf{Figure \ref{fig: contact_dynamics}}, when the beak collides with the plate, penetration into the plate occurs. Based on the resulting penetration depth \(r\), the impedance \(d\) is computed, which in turn determines the contact force.

\section{Additional information of actual robot experiments}

\begin{figure}[H]
\centering
  \includegraphics[width=0.9\linewidth]{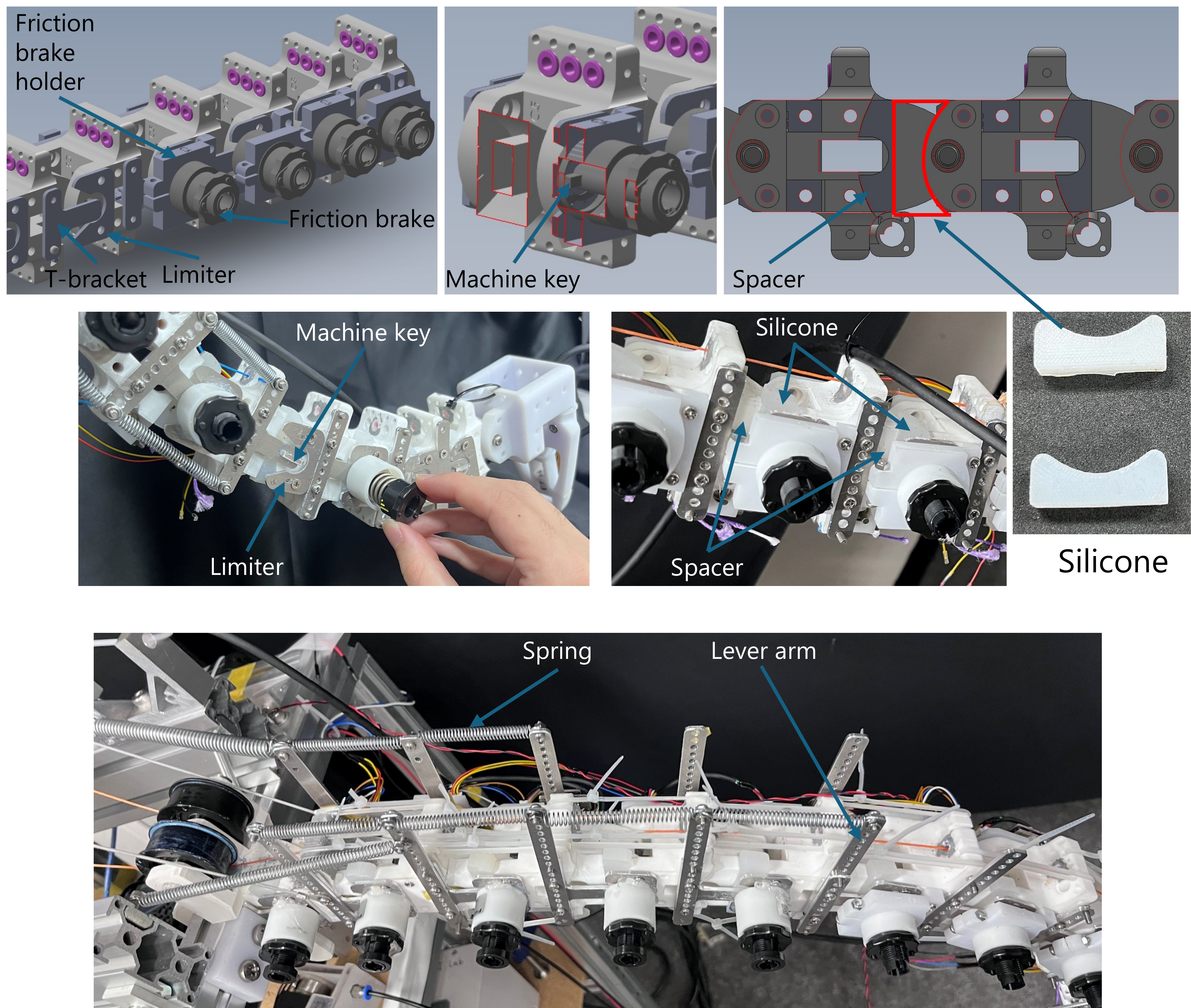}
  \caption{
Detail of RobOstrich
    }
  \label{fig: Detail_of_actual_robot}
\end{figure}

\textbf{Figure \ref{fig: Detail_of_actual_robot}} illustrates the detailed design of each joint in the robot.
Each joint is equipped with a joint angle limiter and a T-bracket.  
Since these components are subject to very high torque, they were made from aluminum alloy using a CNC milling machine. 
To introduce viscosity into the relative motion between the limiter and the T-bracket, we mounted a friction brake on the limiter.
To prevent the brake from detaching due to impact, a 3D printed holder was used.
The friction torque generated by the brake is transmitted to the T-bracket through a machine key.  
However, because there is a delay in generating the friction torque that resists the angular velocity of the joint, the friction brake alone is insufficient to prevent trajectory instability caused by collisions.
To address this, silicone-based materials were inserted into each joint with spacers, providing viscoelastic properties to the joints.
\par The bottom figure in Figure \ref{fig: Detail_of_actual_robot} shows the ligament configuration used for gravity compensation in the RobOstrich.  
Tension springs are installed on the lever arms and are arranged so that the elasticity increases toward the base of the robot.  
Since a large amount of torque is applied near the base, there was a risk of damage due to repeated impacts.
As a result, we did not compare the learning performance of the robot with and without ligaments in the actual robot.

In this paper, Dragon Skin 10 and Ecoflex 00-30 were used as silicone cartilages. 
The mechanical properties of them are shown in Table \ref{table: Mechanical properties}.

\begin{table}[h]
\centering
 \caption{Mechanical properties of silicone for cartilage}
 \label{table: Mechanical properties}
  \begin{tabular}[htbp]{@{}lcccc@{}}
    \hline
                  & Shore Hardness & 100\% Modulus & Mixed Viscosity & Tensile Strength \\
    \hline
    Dragon Skin 10  & 10  &  150 kPa (22 psi) & 23 Pa s (23000 cps) &  3.3 MPa (475 psi)\\
    Ecoflex 00-30   & 30  &  69 kPa (10 psi)  & 3.0 Pa s (3000 cps) &  1.4 MPa (200 psi)\\
    \hline
  \end{tabular}
\end{table}